\documentclass{article}

\usepackage{microtype}
\usepackage{graphicx}
\usepackage{subfigure}
\usepackage{booktabs} %

\usepackage{hyperref}

\usepackage[preprint]{icml2023}

\usepackage{amsmath}
\usepackage{amssymb}
\usepackage{mathtools}
\usepackage{amsthm}

\usepackage[capitalize,noabbrev]{cleveref}

\usepackage{thm-restate}
\theoremstyle{plain}

\crefname{auxlemma}{Aux.\ Lemma}{Aux.\ Lemmas}

\newtheorem{definition}{Definition}
\theoremstyle{definition}
\newtheorem{example}{Example}
\theoremstyle{remark}

\usepackage[framemethod=tikz,splitbottomskip=0cm,xcolor=dvipsnames]{mdframed}
\usetikzlibrary{shapes}
\surroundwithmdframed[outerlinewidth=0pt,
  innerlinewidth=0pt,
  middlelinewidth=0pt,
  middlelinecolor=white,
  innerleftmargin=2pt,
  innerrightmargin=2pt,
  nobreak=false,
  bottomline=true,
  topline=true,
  rightline=true,
  backgroundcolor=GreenYellow!10,
  roundcorner=5pt,
]{example}
\surroundwithmdframed[outerlinewidth=0.0pt,
  innerlinewidth=0.0pt,
  middlelinewidth=0pt,
  middlelinecolor=white,
  innertopmargin=3pt,
  innerleftmargin=3pt,
  innerrightmargin=3pt,
  nobreak=false,
  bottomline=true,
  topline=true,
  rightline=true,
  backgroundcolor=Gray!10,
  roundcorner=3pt,
  startcode={\vspace{1ex}},
]{definition}
\surroundwithmdframed[outerlinewidth=0.0pt,
  innerlinewidth=0.0pt,
  middlelinewidth=0pt,
  middlelinecolor=white,
  innertopmargin=3pt,
  innerleftmargin=3pt,
  innerrightmargin=3pt,
  nobreak=false,
  bottomline=true,
  topline=true,
  rightline=true,
  backgroundcolor=Gray!10,
  roundcorner=3pt,
  startcode={\vspace{1ex}},
]{marktheorem}
\surroundwithmdframed[outerlinewidth=0.0pt,
  innerlinewidth=0.0pt,
  middlelinewidth=0pt,
  middlelinecolor=white,
  innertopmargin=3pt,
  innerleftmargin=3pt,
  innerrightmargin=3pt,
  nobreak=false,
  bottomline=true,
  topline=true,
  rightline=true,
  backgroundcolor=Gray!10,
  roundcorner=3pt,
  startcode={\vspace{1ex}},
]{markproposition}

\usepackage{xspace}
\usepackage{units}
\usepackage[commandRef=cref]{proof-at-the-end}
\usepackage{enumitem}
\usepackage{cuted}
\usepackage{pifont}
\usepackage{afterpage}
\usepackage{printlen}
\uselengthunit{in} 
\usepackage{enumitem}
\setlist{nosep}
\usepackage{pgfkeys}
\usepackage{chngcntr}
\usepackage{wrapfig}

\newcommand{\setvalue}[1]{\pgfkeys{/variables/#1}}
\newcommand{\getvalue}[1]{\pgfkeysvalueof{/variables/#1}}
\newcommand{\declare}[1]{%
 \pgfkeys{
  /variables/#1.is family,
  /variables/#1.unknown/.style = {\pgfkeyscurrentpath/\pgfkeyscurrentname/.initial = ##1}
 }%
}
\declare{}

\setlist[enumerate,1]{label={\roman*.}, noitemsep, topsep=0pt}

\def\RR{\ensuremath{\mathbb{R}}\xspace}
\newcommand{\norm}[1]{\ensuremath{\left\| {#1} \right\|}}
\newcommand{\TODO}[1]{$ $\newline\noindent\colorbox{yellow!30}{\parbox{\dimexpr\the\columnwidth-2\fboxsep}{\textbf{\texttt{TODO:}} \textit{#1}}}}

\def\CalS{\ensuremath{\mathcal{S}}\xspace}
\def\SLip{\CalS-Lipschitz\xspace}
\def\SLipness{\CalS-Lipschitzness\xspace}
\def\Scert{\CalS-certificate\xspace}
\def\Scerts{\CalS-certificates\xspace}
\DeclareMathOperator{\hull}{hull}
\newcommand{\stress}[1]{{\fontfamily{LinuxLibertineT-OsF}\fontseries{sb}\selectfont #1}}
\DeclareMathSymbol{\sm}{\mathbin}{AMSa}{"39}
\def\implies{\Rightarrow}

\def\Ronesymb{\textcolor{teal}{\ding{182}}\xspace}
\def\Rtwosymb{\textcolor{orange}{\ding{183}}\xspace}
\def\Rthreesymb{{\textcolor{purple}{\ding{184}}}\xspace}
\def\ronesymb{\textcolor{teal}{\ding{172}}\xspace}
\def\rtwosymb{\textcolor{orange}{\ding{173}}\xspace}
\def\rthreesymb{\textcolor{purple}{\ding{174}}\xspace}

\def\Rone{\ref{R1}}
\def\Rtwo{\ref{R2}}
\def\Rthree{\ref{R3}}
\def\rone{\ref{r1}}
\def\rtwo{\ref{r2}}
\def\rthree{\ref{r3}}

\newcommand*\introducethenuse[3]{%
  \def#1{#2\renewcommand#1{#3}}%
}

\NewDocumentCommand\wrapword{m}{\tikz[baseline=(X.base)]\node [draw=gray!40,semithick,rectangle,inner sep=0.8pt, rounded corners=2pt] (X) {#1};}

\def\samecasymb{\smash{\wrapword{\strut$c_A^{_{=}}$}}}
\def\sameca{\ref{samecatarget}}
\def\diffcasymb{\smash{\wrapword{\strut$c_A^{_{\neq}}$}\xspace}}
\def\diffca{\ref{diffcatarget}}
\def\samecbsymb{\smash{\wrapword{\strut$c_B^{_{=}}$}\xspace}}
\def\samecb{\ref{samecbtarget}}

\def\ucontsymb{\smash{\wrapword{\textnormal{\textsf{U}}}}}
\def\ucontstarsymb{\smash{\wrapword{\textnormal{\textsf{U}$'$}}}}
\def\cwcontsymb{\smash{\wrapword{\textnormal{\textsf{CW}}}}}
\def\cdcontsymb{\smash{\wrapword{\textnormal{\textsf{CD}}}}}

\introducethenuse\ucont{\hypertarget{uconttarget}{\ucontsymb}}{\hyperlink{uconttarget}{\ucontsymb}\xspace}
\introducethenuse\ucontstar{\hypertarget{ucontstartarget}{\ucontstarsymb}}{\hyperlink{ucontstartarget}{\ucontstarsymb}\xspace}
\introducethenuse\cwcont{\hypertarget{cwconttarget}{\cwcontsymb}}{\hyperlink{cwconttarget}{\cwcontsymb}\xspace}
\introducethenuse\cdcont{\hypertarget{cdconttarget}{\cdcontsymb}}{\hyperlink{cdconttarget}{\cdcontsymb}\xspace}

\definecolor{specialblue}{RGB}{203,100,58}
\def\redpatch{\tikz[baseline=-0.5ex,scale=0.4]{\draw[fill=specialblue,line width=0pt]  circle (1em);}\xspace}
\def\bluepatch{\tikz[baseline=-0.5ex,scale=0.4]{\draw[fill=blue!50,line width=0pt]  circle (1em);}\xspace}

\def\dottedline{\tikz[baseline=-0.ex,scale=0.3]{\draw[dotted,line width=1.2pt](0,0)--(1,1);}\xspace}

\def\solidline{\tikz[baseline=0.2ex,scale=0.3]{\draw[line width=1.2pt](0,0)--(1,1);}\xspace}

\newcommand{\ie}{\textit{i.e.}\xspace}
\newcommand{\eg}{\textit{e.g.}\xspace}

\begin{document} 

\twocolumn[
\icmltitle{Certifying Ensembles: A General Certification Theory with $\mathcal{S}$-Lipschitzness}

\icmlsetsymbol{equal}{*}

\begin{icmlauthorlist}
\icmlauthor{Aleksandar Petrov}{cs,equal}
\icmlauthor{Francisco Eiras}{engsci}
\icmlauthor{Amartya Sanyal}{eth}
\icmlauthor{Philip H.S.\ Torr}{engsci}
\icmlauthor{Adel Bibi}{engsci,equal}
\end{icmlauthorlist}

\icmlaffiliation{cs}{Department of Computer Science, University of Oxford, Oxford, UK}
\icmlaffiliation{engsci}{Department of Engineering Science, University of Oxford, Oxford, UK}
\icmlaffiliation{eth}{ETH AI Center, ETH Z\"urich, Z\"urich, Switzerland}

\icmlcorrespondingauthor{A. Petrov}{aleks@robots.ox.ac.uk}
\icmlcorrespondingauthor{A. Bibi}{adel.bibi@eng.ox.ac.uk}

\icmlkeywords{Machine Learning, ICML, Robustness, Ensemble learning}

\vskip 0.3in
]

\printAffiliationsAndNotice{\icmlEqualContribution} %

\begin{abstract}
   Improving and guaranteeing the robustness of deep learning models has been a topic of intense research.
   Ensembling, which combines several classifiers to provide a better model, has shown to be beneficial for generalisation, uncertainty estimation, calibration, and mitigating the effects of concept drift. 
   However, the impact of ensembling on certified robustness is less well understood. 
   In this work, we generalise Lipschitz continuity by introducing $\mathcal{S}$-Lipschitz classifiers, which we use to analyse the theoretical robustness of ensembles. 
   Our results are precise conditions when ensembles of robust classifiers are more robust than any constituent classifier, as well as conditions when they are less robust. 
\end{abstract}

\section{Introduction}
\label{sec:introduction}

Deep learning classifiers are almost as celebrated for their near-perfect accuracy, as they are notorious for their lack of robustness~\citep{biggio_evasion_2013,szegedy_intriguing_2014,goodfellow_explaining_2015}.
Within the past decade, as empirically robust classifiers have begun to emerge \citep{madry2017towards,wang2018mixtrain}, so did attempts to certify their robustness.
The goal of robustness certification is to obtain a set of additive perturbations around an input %
under which the prediction remains unchanged.
Most approaches fall under one of three families of methods: exact certification \citep{katz2017reluplex,ehlers2017formal,huang2017safety}, over-approximation \citep{wong2018provable,salman2019convex},
or probabilistic certification \cite{weng2019proven}, notably \textit{randomized smoothing} methods~\citep{lecuyer2019certified,cohen_certied_2019}.

Ensembling consists in combining several classifiers to obtain a better-performing one \citep{hansen1990neural,omer2018ensemble}.
While it was originally proposed to improve the accuracy of weak classifiers \citep{rokach2016decision,allen2020towards}, it is also beneficial for improving uncertainty estimation and calibration \citep{Lakshminarayanan2017simple,zhang20MixnMatch}, as well as mitigating the effects of concept drift \citep{omer2018ensemble}. 
These benefits of ensembling have inspired research into studying its effect on robustness.
For example, recent empirical works have shown that encouraging diversity in the non-maximal predictions \citep{pang19improving}, or in the gradient directions \citep{kariyappa2019improving} of individual classifiers results in ensembles with improved robustness.

However, the degree of improved performance depends on the ensembled classifiers. 
When the constituent classifiers are all highly accurate, there is little room for improvement after ensembling; the gains are most pronounced with weak classifiers. %
Possibly, a similar limitation holds for robustness: perhaps ensembles of robust classifiers enjoy lower robustness improvements than ensembles of non-robust classifiers.
\citet{pang19improving}, \citet{horvath2021boosting}, \citet{yang2022on} and \citet{puigcerver2022adversarial} propose theoretical justifications for why ensembles boost robustness but stop short of quantifying the improvement, especially when the individual classifiers are already robust.
This raises the following questions on the robustness limitations of ensembles:
\begin{enumerate}
    \item \emph{For a collection of robust classifiers, can their ensemble be more robust than its constituents? If so, what is the maximum achievable improvement, and under which conditions?}
    \item Conversely: \emph{Is it possible for an ensemble of robust classifiers to be less robust than its constituents? If so, what is the worst possible drop in robustness, and under which conditions?}
\end{enumerate}

We tackle these questions by introducing \SLipness in \cref{sec:single_certificates},  a generalization of Lipschitz continuity that enables tight analysis of the theoretical robustness of ensembles.
\SLipness gives rise to certificates which need not be symmetric and are guaranteed to certify regions at least as large as the classical Lipschitz ones. 

Building on the \SLipness framework, in \cref{sec:smooth_ensembles}, we offer the following answers to the above questions:
\begin{enumerate}
    \item It is possible for ensembles to certify every perturbation that any of the individual classifiers can certify, and even a \emph{superset of their union}. However, we note that the gain is most pronounced when the individual classifiers are not robust; as the robustness of the individual classifiers improves, the robustness gain from ensembling becomes more limited.
    \item It is possible for ensembles to fail to certify perturbations that every single one of the individual classifiers certifies, \eg the ensemble certificate can be a proper \emph{subset of the intersection} of the constituent certificates.
        Interestingly, in the worst case, ensembles of robust classifiers \emph{do not certify any perturbation at all}.
        However, we show that as long as all classifiers have the same prediction, the ensemble certificate will never be a \emph{subset of the intersection}.
\end{enumerate}

\section{Related work}

\stress{Certified Adversarial Robustness.}
Deep neural networks are vulnerable to adversarial attacks~\citep{szegedy_intriguing_2014,goodfellow_explaining_2015}. 
The emergence of empirical defences to these mechanisms ~\citep{papernot2017practical,madry2017towards,de2022make}, has motivated the need for methods that achieve \textit{certified} robustness. 
Those methods can be classified into \textit{exact}, \ie, complete \citep{katz2017reluplex,ehlers2017formal,huang2017safety,lomuscio2017approach,bunel2018unified}, or \textit{conservative}, \ie, sound but incomplete \citep{gowal2018effectiveness,mirman2018differentiable,wang2018mixtrain,ayers2020parot}. 
Probabilistic methods, mostly based on \textit{randomized smoothing}~\citep{lecuyer2019certified,cohen_certied_2019}, have been shown to scale to large networks but have high inference time complexity.

\stress{Robustness of Ensembles.} 
While ensembles have long been used to boost the accuracy of classifiers, interest in their robustness properties is rather recent.
\citet{pang19improving} propose a regulariser that diversifies the non-maximal predictions of individual classifiers which leads to empirically better robustness.
\citet{kariyappa2019improving} recommend a different type of regularisation: \emph{Diversity Training} which encourages misaligned gradients. 
Moreover, \citet{horvath2021boosting} and \citet{yang2022on} observe that applying randomized smoothing after ensembling results in more certifiably robust models than applying it to the individual classifiers.
\citet{xu2021mixture} proposed using a mixture of clean and robust experts, while \citet{puigcerver2022adversarial} studied the Lipschitz continuity of ensembles.

\section{\texorpdfstring{\CalS}{S}-Certificates with \texorpdfstring{\CalS}{S}-Lipschitzness}
\label{sec:single_certificates}

We start by introducing the definition of point-wise adversarial robustness of a classifier\footnote{A list of symbols is provided in \cref{sec:symbols_list}.}. 
\begin{definition}[Robustness]
    Given a classifier $f{:}\ \RR^d \to \RR^K$, an $x\in\RR^d$ and a set $Q \subset \RR^d$, $f$ is said to be \emph{robust} at $x$ if
    $\arg\max_{i\in 1,\ldots,K } f_i(x) = \arg\max_{i\in 1,\ldots,K} f_i(x+\delta), ~\forall \delta \in Q,$
    where $f_i$ is the prediction for the $i$-th class.
    We will call $Q$ a \emph{certificate} at $x$.
\end{definition}

As $Q$, also known as a \emph{perturbation set}, depends on $x$, this notion of robustness is also called \emph{point-wise robustness}. 
We start by reviewing the classical notion of Lipschitzness and its relation to robustness before introducing \SLipness: our generalization that permits more general certificates.

\subsection{Lipschitz Certificates}
\label{sec:lipschitz_cert}

The Lipschitz continuity\footnote{Some works refer to \emph{Lipschitz continuity} as \emph{smoothness}.} of a classifier is linked to its robustness.
The predictions of Lipschitz  classifiers with smaller Lipschitz constant change less for the same input perturbations compared to Lipschitz classifiers with a larger constant.
Hence, Lipschitz continuity is commonly used for robustness analysis of neural networks \cite{hein2017formal,bartlett2017spectrally,cisse2017parseval,weng2018evaluating,huang2021training,zhang2021towards,zhang2022rethinking,eiras2022ancer,alfarra22data,Alfarra2022DeformRS}.

The Lipschitz constant of a function is closely related to its gradients.
The larger the norm of the gradients, the more sensitive the function is to perturbations and the larger its Lipschitz constant becomes.
Furthermore, given a Lipschitz classifier with a Lipschitz constant $L$, the \emph{prediction gaps}, \ie, the differences between the confidence of the top prediction and the other classes, fully determine the certificate $Q$.
As such, we have the following proposition.

\begin{theoremEnd}[all end]{auxlemma}
    \label{thm:lip_robustness}
    Consider a classifier $f: \RR^d \to \RR^K$ such that $f_i$ is $L_i$-Lipschitz with respect to the norm $\norm{\cdot}$, that is $|f_i(x)-f_i(x')| \leq L_i \norm{ x - x' }$, $\forall i, x, x'$.
    Then, at a fixed $x$, we have $\arg\max_i f_i(x+\delta)=c_A$ for all $\norm{ \delta } \leq \min_{i\neq c_A} \nicefrac{\left( f_{c_A}(x) - f_i(x) \right)}{\left( L_{c_A} + L_{i} \right)} $, where $c_A = \arg\max_i f_i(x)$.
\end{theoremEnd}
\begin{proofEnd}

    From the definition of $f_i$ being $L_i$-Lipschitz it follows that for all $i=1,\ldots,K$:
    \begin{align*}
        f_i (x) + L_{i}\norm{\delta_i} \geq f_i(x+\delta_i) &\geq f_i (x) - L_{i}\norm{\delta_i}
    \end{align*}
    For $\arg\max_c f_c(x+\delta) = c_A$ it must be that $f_i(x+\delta) \leq f_{c_A}(x+\delta)$ for all $i\neq c_A$. By applying the above inequalities for every $i\neq c_A$ we obtain:
    \begin{align}
    f_{c_A} (x) - L_{c_A}\norm{\delta_i} - f_i (x) - L_{i}\norm{\delta_i} & \geq 0 \nonumber \\
    \norm{\delta_i} \leq \frac{f_{c_A}(x) - f_i(x)}{L_{c_A} + L_{i}}. \label{eq:lip_robustness_one_class_bound}
    \end{align}
    \Cref{eq:lip_robustness_one_class_bound} is an upper bound of the perturbation that will not change the prediction from $c_A$ to $i$.
    Since this must hold for all $i\neq c_A$, it is only valid for the intersection of these perturbation sets, \ie, $\norm{\delta} \leq \min_{i\neq c_A} \nicefrac{(f_{c_A}(x) - f_i(x))}{(L_{c_A} + L_{i})}$.
\end{proofEnd}
\begin{theoremEnd}[all end]{auxlemma}
    \label{thm:gradients_lipschitz}
    Consider a differentiable $h:\RR^d \to \RR$, such that $\sup_x \norm{\nabla h (x)}_\star \leq L$, where $\norm{\cdot}_\star$ is the dual norm of $\norm{\cdot}$.
    Then $h$ is $L$-Lipschitz with respect to $\norm{\cdot}$.
\end{theoremEnd}
\begin{proofEnd}
    See proof of Proposition 1 from~\cite{eiras2022ancer}.
\end{proofEnd}

\stepcounter{footnote}
\pgfmathparse{int(\value{footnote})}
\setvalue{differentiability_footnote = \pgfmathresult}
\footnotetext[\value{footnote}]{For simplicity, we work with differentiable classifiers, even though our results are also valid for continuous classifiers that are not differentiable at finite number of points.
}

\begin{theoremEnd}[end, restate]{proposition}[Certification of Lipschitz classifiers]
    \label{thm:bounded_gradients_certificates}
    Take a differentiable\footnotemark[\getvalue{differentiability_footnote}] classifier $f:\RR^d \to \RR^K$ such that $\sup_x \norm{\nabla f_i (x)}_\star \leq L_i$, $\forall i$.
    Then $f_i$ is $L_i$-Lipschitz with respect to $\|{\cdot}\|$.
    Moreover, $f$ has a certificate %
    \begin{equation}
        \resizebox{\columnwidth}{!}{
            $\displaystyle{Q {=} \left\{\delta\in\RR^d : \norm{ \delta } {\leq} \min_{i\neq c_A} \frac{ f_{c_A}(x) {-} f_i(x) }{L_{i}{+}L_{c_A}} {=} \min_{i\neq c_A} \frac{r_i}{L_{i}{+}L_{c_A}} \right\} }.$
        }
        \label{eq:cwlipschitz}
    \end{equation}
    Here, $\|{\cdot}\|_\star$ is the dual norm to $\|{\cdot}\|$ and $c_A$ is $\arg\max_i f_i(x)$. %
    If all classes have the same Lipschitz constant $L$, \ie, $L_i\leq L, \forall i$, the certificate simplifies to
    \begin{equation}
        Q = \left\{\delta\in\RR^d :\norm{ \delta } \leq \frac{f_{c_A}(x) - f_{c_B}(x) }{2 L} = \frac{r_{c_B}}{2L} \right\},
        \label{eq:ulipschitz}
    \end{equation}
    where $c_B = \arg\max_{i\neq c_A} f_i(x)$.
\end{theoremEnd}
\begin{proofEnd}
    Follows directly from \cref{thm:lip_robustness,thm:gradients_lipschitz}.
\end{proofEnd}

We refer to the formulation in \cref{eq:cwlipschitz} as \emph{class-wise Lipschitz continuity} (\cwcont) since it accounts for the classes potentially having different Lipschitz constants.
Often, however, in prior art, all classes are considered to have the same Lipschitz constant $L$ set such that $L\geq \max_i L_i$.
We refer to this setting captured by \cref{eq:ulipschitz} as \emph{uniform Lipschitz continuity} (\ucont).
Moreover, the Lipschitz certificates apply to any choice of norm; the main text considers only $\ell_p$ norms but we give further examples in \Cref{sec:additional_lipschitz_examples}.

\begin{example}[$\ell_p$ certificates]
    \label{ex:lp_certificates}
    We can construct $\ell_p$ Lipschitz certificates, by bounding the supremum of the dual $\ell_q$ 
    norm of the classifier gradients, where $\nicefrac{1}{p}+\nicefrac{1}{q}=1$.
    This follows directly from H\"{o}lder's inequality.
\end{example}

\begin{figure}
    \centering
    \includegraphics[width=0.95\columnwidth]{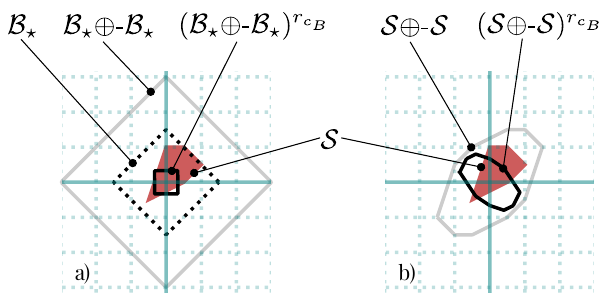}
    \caption{
        Lipschitz certificate for the $\ell_\infty$ norm (a) and \Scert (b).
        These are shown with \solidline.
        The \Scert is a superset of the Lipschitz certificate.
        Both certificates are in the uniform setting (\ucont) for a classifier $f:\RR^d{\to}\RR^K$ with range space of gradients $\CalS=\{\nabla f_i(x) : x\in\RR^d, i{=}1,\ldots,K\}$ (shown in \redpatch). 
        We assume $r_{c_B} = 1$.
        $\mathcal B_\star$ is the smallest $\ell_1$ ball containing $\CalS$.
    }
    \label{fig:linf_vs_S_cert}
\end{figure}

\cref{fig:linf_vs_S_cert}a demonstrates the intimate relationship between the norm of the gradients of a classifier, \ie, its Lipschitzness, and the resulting certificates from \cref{thm:bounded_gradients_certificates}. 
Take a classifier $f:\RR^d \to \RR^K$ and the set of all its gradients $\CalS{=}\{\nabla f_i(x) : x\in\RR^d, i{=}1,{\ldots},K\}$ shown in \redpatch.
For simplicity, assume also that $r_{c_B}{=}1$.
As $\sup_{s\in\CalS} \|s\|_1 {\leq} 1.5$, the $f_i$ are $1.5$-Lipschitz with respect to the $\ell_\infty$ norm.
Therefore, from \cref{eq:ulipschitz} the certificate $Q$ is the $\ell_\infty$ ball of radius $\nicefrac{1}{3}$ shown with \solidline. 
Taking the supremum of the $\ell_1$ norm introduces overapproximation of the true set of gradients.
Note how the \dottedline\ region has the same supremum $\ell_1$ norm as $\CalS$ and hence has the same certificate \solidline.
However, \dottedline\ is a superset of the gradients \CalS and must correspond to a more sensitive classifier.
This is due to the overapproximating action of the supremum of the gradient norms.
To rectify this, we offer a novel generalization of Lipschitzness working directly with the gradients $\CalS$.

\begin{figure*}
    \centering
    \includegraphics[width=\textwidth]{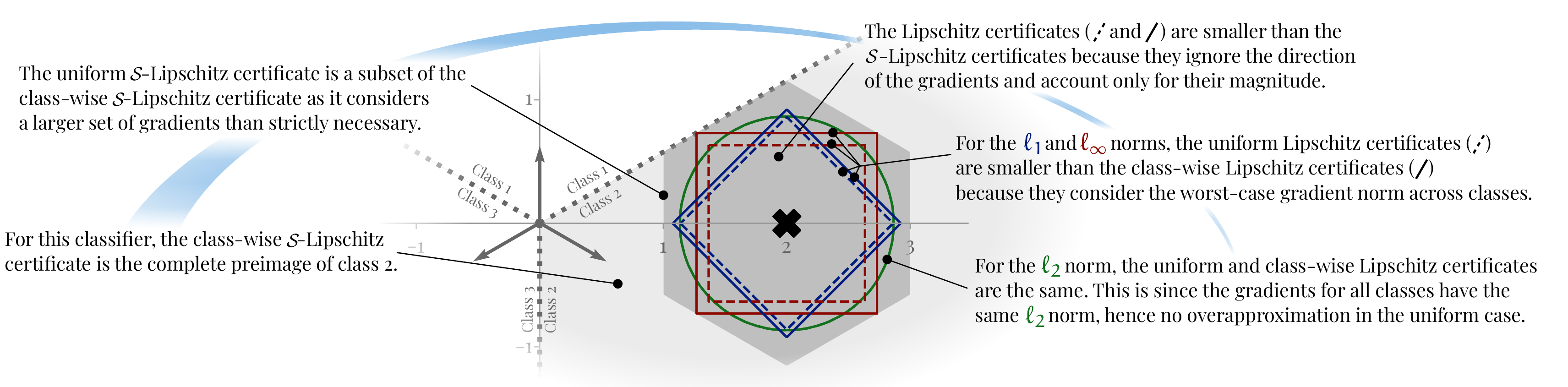}
    \caption{Lipschitz and \SLip certificates at $x=[2,0]^\top$ for a linear classifier that splits the domain into three equal sectors. Step-by-step explanation of the construction of the certificates is provided in \cref{sec:2d_linear_classifier_workedout}.}
    \label{fig:2d_linear_classifier}
\end{figure*}

\subsection{\texorpdfstring{\CalS}{S}-Certificates}
\label{sec:Slipschitz_cert}

We observed that Lipschitzness induces a larger gradient overapproximation to the set of gradients set $\CalS$. 
This begs the question: \emph{Can we enlarge the certificates by avoiding the dual norm ball overapproximation of the gradients and work directly with the exact gradient set $\CalS$?}

To this end, we first generalize the definition of a Lipschitz function which allows the use of the exact range space of the gradient as opposed to any overapproximation.

\begin{definition}[\SLip function]
    \label{def:Slip}
    A function $f:\RR^d \to \RR$ is \SLip for a bounded set $\CalS\subset\RR^d$ if it holds that:
    $$ -\rho_\CalS (x-y) \leq f(y)-f(x) \leq \rho_\CalS (y-x), ~ \forall x,y \in \RR^d,$$
    with $\rho_\CalS(\delta)=\sup_{c\in\CalS} c^\top \delta$.
    If \CalS is convex, then $\rho_\CalS$ corresponds to its support function.
\end{definition}

Intuitively, $\rho_\CalS(\delta)$ is the biggest change in direction $\delta$ that we can incur using the gradients in $\CalS$.
Note that the \SLipness generalizes the previous definition of a Lipschitz function. To see this, consider the case where $\CalS=\{x : \norm{x}_\star \leq L\}$. 
Following H\"{o}lder's inequality, we observe that Definition \ref{def:Slip} reduces to the classical $L$-Lipschitzness definition with respect to $\norm{\cdot}$ norm.

In contrast to the classical Lipschitzness, \SLipness accounts not only for the magnitude of the gradients but also for their direction.
We also can generalize the notion of dual norms to sets that are not norm balls:
\stepcounter{footnote}
\pgfmathparse{int(\value{footnote})}
\setvalue{polarset_footnote = \pgfmathresult}
\footnotetext[\value{footnote}]{We are extending the 
    standard notion of a polar set \citep{rockafellar1970convex} to encompass radii different from 1.}
\begin{definition}[Polar set]
    \label{def:polar_set}
    For a set $\CalS\subset\RR^d$, the polar set\footnotemark[\getvalue{polarset_footnote}] to \CalS\ of radius $r > 0$ is defined as:
    $$ (\CalS)^r = \left\{ \delta \in\RR^d ~:~ \rho_\CalS(\delta) = \textstyle{\sup_{x\in\CalS}} x^\top \delta \leq r \right\}.$$ 
\end{definition}

Take $f:\RR^d \to \RR$ to be \SLip with $\CalS = \{ x\in\RR^d : \norm{x}_1 \leq L \}$. 
Then, the polar set $(\CalS)^r$ of radius $r$ is the perturbation set that will not change $f$ by more than $r$.
$(\CalS)^r$ is $\{ \delta \in\RR^d~:~\norm{\delta}_\infty \leq \nicefrac{r}{L} \}$ which is the same result that follows from $f$ being $L$-Lipschitz.
We are now ready to generalize \Cref{thm:bounded_gradients_certificates} with \SLipness:
\begin{theoremEnd}[all end, restate]{auxlemma}
    \label{thm:hull_can_be_dropped}
    For a bounded set $\CalS\subseteq\RR^d$, it holds that $\rho_{\hull\CalS} (\delta) = \rho_\CalS (\delta),~ \forall \delta\in\RR^d.$
    In other words, $f$ being \SLip is the same as it being $(\hull\CalS)$-Lipschitz.
\end{theoremEnd}
\begin{theoremEnd}[all end]{auxlemma}[\SLip function and gradients]
    \label{thm:gradients_to_S}
    Consider a differentiable $f:\RR^d\to\RR$. If $\nabla f:\RR^d\to\CalS$, then $f$ is \SLip.
    The reverse also holds: if $f$ is \SLip, then, $\nabla f(x) \in \hull\CalS, ~\forall x\in\RR^d$.
\end{theoremEnd}
\begin{proofEnd}
    Let's first start by showing that a function with bounded gradients is \SLip.
    Consider any $x,y\in\RR^d$ and define $\gamma:[0,1]\to\RR^d$ where $\gamma(t)=(1-t)x+ty$. 
    Then we have that:
    \begin{align*}
    f(y)-f(x) &= f(\gamma(1))-f(\gamma(0)) \\
              &= \int_0^1 \frac{df(\gamma(t))}{dt} dt \\
              &= \int_0^1 \frac{df(\gamma(t))}{d\gamma(t)} \frac{d\gamma(t)}{dt} dt \\
              &= \int_0^1 \nabla_x f(\gamma(t))^\top \nabla_t \gamma(t) dt \\
              &= \int_0^1 \nabla_x f((1-t)x+ty)^\top (y-x) dt \\
              &\leq \int_0^1 \max_{t\in[0,1]} \left\{ \nabla_x f((1-t)x+ty)^\top (y-x) \right\} dt \\
              &= \max_{t\in[0,1]} \left\{ \nabla_x f((1-t)x+ty)^\top (y-x) \right\} \\
              &\leq \sup_{\nabla f \in\CalS} \nabla f^\top (y-x) \\
              &= \rho_\CalS(y-x),
    \end{align*}
    where we used the fundamental theorem of calculus, the fact that $f$ is continuous, and \cref{def:Slip}.
    Similarly,
    \begin{align*}
    f(y)-f(x) &\geq \int_0^1 \min_{t\in[0,1]} \left\{ \nabla_x f((1-t)x+ty)^\top (y-x) \right\} dt \\
              &= \min_{t\in[0,1]} \left\{ \nabla_x f((1-t)x+ty)^\top (y-x) \right\} \\
              &= -\max_{t\in[0,1]} \left\{ \nabla_x f((1-t)x+ty)^\top (x-y) \right\} \\
              &\geq -\sup_{\nabla f \in\CalS} \nabla f^\top (x-y) \\
              &= -\rho_\CalS(x-y).
    \end{align*}

    Next, let's show the reverse: that if a function is \SLip, then its gradients must be in $\hull\CalS$.
    If $f$ is \SLip, then $f(y)-f(x)\leq\sup_{c\in\CalS} c^\top (y-x)$.
    Consider the directional derivative of $f$ in direction $v\in\RR^d$ at $x$, and taking $y=x+hv$:
    \begin{align*}
        \nabla_v f(x) 
        &= \lim_{h\to 0} \frac{f(x+hv)-f(x)}{h} \\
        &\leq \lim_{h\to 0} \frac{\sup_{c\in S} c^\top(hv)}{h} \\
        &= \lim_{h\to 0} \frac{h\sup_{c\in S} c^\top v}{h} \\
        &\overset{\ast}{=} \sup_{c\in S} c^\top v \\
        &= \rho_\CalS (v),
    \end{align*}
    with $\ast$ following from the L'H\^opital's rule.
    Similarly, $\nabla_v f(x) \geq -\rho_\CalS (-v),~ \forall x\in\RR^d$.
    Hence, we have:
    \begin{equation}
       -\rho_\CalS (-v) = - \sup_{c\in S} c^\top (-v) \leq  \nabla f(x)^\top v \leq \sup_{c\in S} c^\top v .
       \label{eq:S_to_gradients_bounds}
    \end{equation}
    Now we need to show that \Cref{eq:S_to_gradients_bounds} implies that $\nabla f(x)\in \hull\CalS$.
    By the properties of support functions of convex sets we have that $\nabla f(x)\in\hull\CalS$ iff
    $$ \sup_{c'\in\hull\CalS} \left\{ c'^\top \nabla f(x) - \rho_\CalS (c')\right\} = 0.$$
    In the above, we use that $\rho_\CalS (\delta) = \rho_{\hull\CalS}(\delta)$ (\cref{thm:hull_can_be_dropped}).
    Substituting from \Cref{eq:S_to_gradients_bounds} we get:
    \begin{equation*}
        \begin{split}
            \sup_{c'\in\hull\CalS} &\left\{ c'^\top \nabla f(x) - \rho_\CalS (c')\right\} \\
            &\leq \sup_{c'\in\hull\CalS} \left\{ \sup_{c\in S} c^\top c' - \sup_{c\in\CalS} c^\top c' \right\} = 0,
        \end{split}
    \end{equation*}
    hence, $\nabla f(x) \in \hull\CalS, ~\forall x\in\RR^d$.
\end{proofEnd}

\begin{theoremEnd}[all end]{auxlemma}
    \label{lemma:rho_negation}
    Given $\CalS\subseteq\RR^n$, it holds that $\rho_\CalS (\delta) = \rho_{-\CalS}(-\delta), ~ \forall \delta\in\RR^d$.
\end{theoremEnd}
\begin{proofEnd}
    $\rho_{-\CalS}(-\delta) = \sup_{c\in-\CalS} c^\top (-\delta) = \sup_{c\in\CalS} (-c)^\top(-\delta) = \sup_{c\in\CalS} c^\top \delta = \rho_\CalS (\delta).$
\end{proofEnd}

\begin{theoremEnd}[all end]{auxlemma}
    \label{lemma:rho_sum_with_neg}
    Given $\CalS,\CalS'\subseteq\RR^n$, for all $\delta\in\RR^d$ it holds that $\rho_\CalS(\delta) + \rho_{-\CalS'}(\delta) = \rho_{\CalS\oplus-\CalS'}$, where $\oplus$ is the Minkowski sum operator.
\end{theoremEnd}
\begin{proofEnd}
    \begin{align*}
        \sup_{c\in\CalS} (c^\top \delta) + \sup_{c'\in-\CalS'} (c'^\top \delta)
        &= \sup_{c\in\CalS} (c^\top \delta) + \sup_{c'\in\CalS'} (-c'^\top \delta) \\
        &= \sup_{c\in\CalS,c'\in\CalS'} (c-c')^\top \delta.
    \end{align*}
    At the same time, by definition of the Minkowski sum:
    $$ \rho_{\CalS\oplus-\CalS'} (\delta) 
    = \sup_{c\in\CalS\oplus-\CalS'} c^\top \delta 
    = \sup_{c\in\CalS,c'\in-\CalS'} (c+c')^\top \delta .$$
\end{proofEnd}
\begin{theoremEnd}[end, restate]{theorem}[\Scerts]
    \label{thm:Slip_certificate}
    Let $f:\RR^d\to\RR^K$ be a classifier with $f_i$ being differentiable and $\nabla f_i: \RR^d\to\CalS_i$ for all $i=1,\ldots,K$.
    Then, each $f_i$ is $\CalS_i$-Lipschitz.
    Furthermore, for a fixed $x$, $f$ is robust at $x$ against all $\delta$ in
    \begin{equation}
        Q = \textstyle{\bigcap_{i\neq c_A}} \left( \CalS_i \oplus - \CalS_{c_A} \right) ^{r_i}.
        \label{eq:Slip_certificate_cw}
    \end{equation}
    Here, $c_A {=} \arg\max_c f_c(x)$, $r_i{=}f_{c_A}(x) {-} f_i (x)$, and $\oplus$ is the Minkowski sum.
    If $\CalS \supseteq \CalS_i, \forall i$, then we have the simplified certificate
    \begin{equation}
        Q = (\CalS\oplus -\CalS)^{r_{c_B}},
        \label{eq:Slip_certificate_uniform}
    \end{equation}
    where $c_B = \arg\max_{c\neq c_A} f_c(x)$.%
\end{theoremEnd}
\begin{proofEnd}
    The connection between gradients and \SLipness comes from \cref{thm:gradients_to_S}.
    
    Following \Cref{def:Slip}, we have:
    \begin{align*}
        f_{c_A} - \rho_{\CalS_{c_A}} (x-y) &\leq f_{c_A}(y) \\
        f_i + \rho_{\CalS_{i}} (y-x) &\geq f_i(y), ~\forall i\neq c_A.
    \end{align*}
    We want $f_{c_A}(y)>f_i(y), ~\forall i\neq c_A$, hence, a sufficient condition following the two inequalities above is
    \begin{equation*}
        \begin{split}
            f_i + \rho_{\CalS_{i}} (y-x) &< f_{c_A} - \rho_{\CalS_{c_A}} (x-y) \iff \\
            &\rho_{\CalS_{i}} (y-x) + \rho_{\CalS_{c_A}} (x-y) <  f_{c_A} - f_i.
        \end{split}
    \end{equation*}
    Using \Cref{lemma:rho_negation,lemma:rho_sum_with_neg} and setting $\delta=y-x$ we get:
    $$ \rho_{\CalS_{i}\oplus-\CalS_{c_A}} (\delta) < f_{c_A} - f_i,$$
    which is the definition of $(\CalS_{i}\oplus-\CalS_{c_A})^{r_i}$.
    As this needs to hold for all $i\neq c_A$ we take the intersection.

    To show \cref{eq:Slip_certificate_uniform} observe that ${r_{c_B}}{\leq} r_i, \forall i\neq c_A$.
    Hence, by \Cref{thm:polar_set_properties}iii, $\left(\CalS_i\oplus -\CalS_{c_A} \right)^{r_i} \supseteq \left(\CalS_i\oplus -\CalS_{c_A} \right)^{r_{c_B}} $.
    The rest follows from $\CalS_i \subseteq \CalS$ and \cref{thm:polar_set_properties}iv.
\end{proofEnd}

Note the similarities between \cref{thm:bounded_gradients_certificates,thm:Slip_certificate}.
$\CalS_i$ generalizes the Lipschitz constant $L_i$, while the polar set generalizes the dual norm.
$(\CalS_i \oplus - \CalS_{c_A})^{r_i}$ is the certificate that the prediction does not change from $c_A$ to $i$.
Taking the intersection in \cref{eq:Slip_certificate_cw} ensures that $c_A$ will not be mistaken for any other class.
This corresponds to the $\min$ in \cref{eq:cwlipschitz}.
We also have the \cwcont\ (\cref{eq:Slip_certificate_cw}) and \ucont\ (\cref{eq:Slip_certificate_uniform}) modes, mapping to the same modes for the Lipschitz case (\cref{eq:cwlipschitz,eq:ulipschitz}).
Furthermore, we show \Cref{thm:Slip_certificate} is tight in an example in \Cref{thm:Slip_certificate_tight}. 

\begin{theoremEnd}[all end]{proposition}%
[Tightness of \Scerts]
    \label{thm:Slip_certificate_tight}
    For any $\delta\not\in(\CalS\oplus-\CalS)^r$, there exists an $f:\RR^d\to\RR^K$ with $f_i$ $\CalS$-Lipschitz for all $i$ and $r_{c_B}=f_{c_A}(x)-f_{c_B}(x)$ such that  $\arg\max_i f_i(x+\delta) \neq c_A$.
\end{theoremEnd}
\begin{proofEnd}
    Let's take a constructive approach and provide a classifier that classifies $x$ and $x+\delta$ differently.
    For simplicity, we will consider a binary classifier.
    Let's fix $x \in\RR^d$, $\delta\not\in(\CalS\oplus-\CalS)^{r_{c_B}}$ and construct a classifier that reduces the gap between $c_A$ and $c_B$ as much as possible, while still being \SLip.
    Take $c, c'\in\CalS$ that attain the supremum
    $$\rho_{\CalS\oplus-\CalS}(\delta) 
    = \sup_{c\in\CalS,~c'\in\CalS} (c-c')^\top \delta 
    > r_{c_B}.$$
    Note that $c, c'$ depend only on $\CalS$ and $\delta$ but not on the classifier $f$.
    Now, let's define the classifier $f$ as:
    \begin{align*}
        f_{c_A}(y) &= (y-x)^\top c' +r_{c_B} \\
        f_{c_B}(y) &= (y-x)^\top c.
    \end{align*}
    We can verify that $f_{c_A}(x) > f_{c_B}(x)$ and that $f_{c_A}(x) - f_{c_B}(x) = r_{c_B}$, as well as that $f_{c_A}$ and $f_{c_B}$ are $\CalS$-Lipschitz, hence satisfying all requirements.
    However, we also have:
    \begin{align*}
          &f_{c_A}(x+\delta) - f_{c_B}(x+\delta)\\
        =\ &\delta^\top c' + r_{c_B} - \delta^\top c = r_{c_B} - (c-c')^\top\delta \\
        <\ &0,
    \end{align*}
    hence $\arg\max_i f_i(x+\delta) = c_B$.
\end{proofEnd}

The certificate in \cref{thm:Slip_certificate} is a polar set (or intersection of polar sets), hence, it has a natural dependence on the gradient sets $\CalS$ and the prediction gap $r$:%

\begin{theoremEnd}[end, restate]{proposition}[Polar set dependence on $\CalS$ and $r$]
    \label{thm:polar_set_properties}
    Let $\CalS,\CalS_1,\CalS_2,\CalS_3,\CalS_4\subset\RR^d$ be bounded and $r,r_1,r_2 > 0$:
    \begin{enumerate}
        \item $\CalS_1 \subseteq \CalS_2 \implies (\CalS_1\oplus\sm\CalS_1) \subseteq (\CalS_2\oplus\sm\CalS_2)$; 
        \item $\CalS_1 \subseteq \CalS_2 \implies (\CalS_1)^r \supseteq (\CalS_2)^r$;
        \item $r_1 \leq r_2 \implies (\CalS)^{r_1} \subseteq (\CalS)^{r_2}$;
        \item \resizebox{0.92\columnwidth}{!}{$((\CalS_1{\subseteq}\CalS_3) \land (\CalS_2{\subseteq}\CalS_4)) \implies (\CalS_3{\oplus}{\sm}\CalS_4)^r \subseteq (\CalS_1{\oplus}{\sm}\CalS_2)^r$.}
    \end{enumerate}
    where $\oplus$ is the Minkowski sum operator.
\end{theoremEnd}
\begin{proofEnd}
    \hfill\\ \textit{Proof of i.:}
    For all $s\in(\CalS_1\oplus-\CalS_1)$ there must be some $s',s''\in\CalS_1$ such that $s'-s''=s$.
    But $s',s''\in\CalS_2$ and hence $s'-s''$ must also be in $\CalS_2\oplus-\CalS_2$.
    \hfill\\ \textit{Proof of ii.:}
    We have to show that $\forall y\in\RR^d$ we have $\sup_{x\in\CalS_2} x^\top y \leq r$ implying  $\sup_{x\in\CalS_1} x^\top y \leq r$.
    This is equivalent to showing that 
    \begin{equation}
        \sup_{x\in\CalS_1} x^\top y \leq \sup_{x\in\CalS_2} x^\top y, ~\forall y\in\RR^d.
        \label{eq:polar_set_properties_condii}
    \end{equation}
    We can rewrite the right-hand side as
    \begin{equation*}
        \sup_{x\in\CalS_2} x^\top y = \max\left\{ \sup_{x\in\CalS_1} x^\top y, ~ \sup_{x\in\CalS_2\backslash\CalS_1} x^\top y \right\},
    \end{equation*}
    for all $y\in\RR^d$, hence \cref{eq:polar_set_properties_condii} is always true.
    \hfill\\ \textit{Proof of iii.:}
    If $y\in(S)^{r_1}$ then: $$\sup_{x\in\CalS} x^\top y \leq r_1.$$
    But then it also holds that $\sup_{x\in\CalS2} x^\top y \leq r_2$ as $r_2\geq r_1$ and hence $y\in(\CalS)^{r_2}$.
    \hfill\\ \textit{Proof of iv.:}
    If $y\in (\CalS_3\oplus-\CalS_4)^r$ then for all $s_3\in \CalS_3, s_4 \in \CalS_4$ it holds that $(s_3-s_4)^\top y \leq r$.
    However, as $\CalS_1$ and $\CalS_2$ are subsets of respectively $\CalS_3$ and $\CalS_4$ it must then also hold that $\forall s_1\in\CalS_1, \forall s_2\in\CalS_2$ we have $(s_1-s_2)^\top y \leq r$.
    This implies that $y\in(\CalS_1\oplus-\CalS_2)^r$.
\end{proofEnd}

The statements \textit{i} and \textit{ii} imply that enlarging the set \CalS of an \SLip classifier reduces the certificate $Q$. 
This is since a larger set of possible derivatives means a more sensitive classifier, hence the set of perturbations that would not change the classification is more restricted. 
Similarly, reducing the prediction gap $r$ means that the certificate must be smaller in order to prevent a change of prediction (statement \textit{iii}). 
Statement \textit{iv} implies that any overapproximation to both $\mathcal{S}_1$ and $\mathcal{S}_2$ for a fixed $r$ results in a smaller certificate. %

\subsection{\texorpdfstring{\CalS}{S}-Certificates Subsume Lipschitz Certificates}
\label{sec:SLip_larger_than_Lip}

We introduced \cref{thm:Slip_certificate} in order to avoid overapproximating the gradients of the classifier with a norm ball in the hopes of obtaining larger certificates. 
Figure \Cref{fig:linf_vs_S_cert} compares the Lipschitz and \Scerts and shows that this is indeed the case.
In \cref{sec:lipschitz_cert} we showed that the illustrated classifier is 1.5-Lipschitz with respect to $\ell_\infty$ norm and that its Lipschitz certificate is therefore the $\ell_\infty$ ball of radius $\nicefrac{1}{3}$.
The same result can be viewed as a special case of $\CalS$-certification when we observe that the classifier is $\mathcal{B}_\star$-Lipschitz with $\mathcal{B}_\star = \{x \in \mathbb{R}^d : \|x\|_1 \leq 1.5\}$. 
Hence, for $r_{c_B}=1$, from \cref{eq:Slip_certificate_uniform} we get the same certificate $(\mathcal{B}_\star \oplus \sm\mathcal{B}_\star)^1= (2\mathcal{B}_\star)^1 = \{\delta \in \mathbb{R}^d : \|\delta\|_\infty \leq \nicefrac{1}{3}\}$ (\solidline in \cref{fig:linf_vs_S_cert}a). 
However, if we do not overapproximate $\CalS$ with $\mathcal B_\star$, then \Cref{eq:Slip_certificate_uniform} gives us the \Scert $(\CalS \oplus \sm \CalS)^1$ (\solidline in \cref{fig:linf_vs_S_cert}b).
Clearly, the \Scert is larger than the Lipschitz one. 
\Cref{thm:scert_always_larger_than_lip} in the appendix shows that this is always the case. 
We now address two questions related to the properties of \Scerts.

\begin{theoremEnd}[all end]{proposition}[The \Scert subsumes any Lipschitz certificate]
    \label{thm:scert_always_larger_than_lip}
    Take $f: \RR^d \to \RR^K$ to be a classifier that such that:
    \begin{enumerate}
        \item $f_i$ is $\CalS_i$-Lipschitz for every $i=1,\ldots, K$ and $\CalS_i$ is the smallest such set (\cwcont\ case); or
        \item $f_i$ is $\CalS$-Lipschitz for all $i=1,\ldots, K$ (\ucont\ case) and $\CalS$ is the smallest such set.
    \end{enumerate}
    Consider a fixed input $x\in\RR^d$.
    Then, the corresponding \Scert\ from \cref{thm:Slip_certificate} at $x$ is always a superset of the Lipschitz certificate for any norm $\|\cdot\|$.
\end{theoremEnd}
\begin{proofEnd}
    We will only consider the \cwcont\ case as the \ucont\ follows trivially from it.
    As discussed in the main text, if $f_i$ is $L_i$-Lipschitz with respect to the norm $\|\cdot\|$, then the Lipschitz certificate at $x$ is equal to the $\mathcal B_{i,\star}$-Lipschitz certificate, where $\mathcal B_{i,\star} = \{ y\in\RR^d : \|y\|_\star \leq L_i\}$.
    Formally:
    \begin{align*}
        Q_\text{Lip} &= \left\{\delta\in\RR^d : \norm{ \delta } \leq \min_{i\neq c_A} \frac{r_i}{L_{i}{+}L_{c_A}} \right\}  \\
        &= \bigcap_{i\neq c_A} (\mathcal B_{i,\star} \oplus \sm \mathcal B_{c_A, \star})^{r_i}.
    \end{align*}

    At the same time, the \Scert is:
    $$ Q_\CalS = \bigcap_{i\neq c_A} (\CalS_i\oplus-\CalS_{c_A})^{r_i}. $$
    
    Next, note that $\CalS_i \subseteq \mathcal B_{i, \star}$, regardless of the choice of the norm $\|\cdot\|$.
    This follows from the definitions of $\CalS_i = \{ \nabla f_i(z) : z\in\RR^d\}$ and $\mathcal B_{i,\star} = \{ y\in\RR^d : \|y\|_\star \leq \sup_z \| \nabla f_i(z)\|_\star \}$.

    As $\CalS_i \subseteq \mathcal B_{i, \star}$, from \Cref{thm:polar_set_properties}iv we have:
    $$ (\CalS_i\oplus-\CalS_{c_A})^{r_i} \supseteq  (\mathcal B_{i,\star} \oplus-\mathcal B_{i,\star})^{r_i}.$$
    Finally, as set intersection preserves the superset relation, we have that
    $$ Q_\CalS = \bigcap_{i\neq c_A} (\CalS_i\oplus\sm\CalS_{c_A})^{r_i} \supseteq \bigcap_{i\neq c_A} (\mathcal B_{i,\star} {\oplus} \sm \mathcal B_{c_A, \star})^{r_i} = Q_\text{Lip}. $$
    
\end{proofEnd}

\stress{Could it be that the \Scert in \Cref{fig:linf_vs_S_cert} is larger than the Lipschitz certificate because of a suboptimal choice of norm?}
No, because whenever the set of gradients is not centrally symmetric, \ie, $\mathcal{S} \neq -\mathcal{S}$, then no matter what norm we choose, we have $\mathcal{B}_\star \supset \CalS$ and thus an \Scert larger than the Lipschitz certificate.
This is because norms are centrally symmetric by definition.

\stress{Are \cwcont\ certificates always supersets to the \ucont\ certificates?} 
The \cwcont\ and \ucont\ \Scerts are larger than any Lipschitz certificate (\cref{thm:scert_always_larger_than_lip}).
As \cwcont\ generalizes \ucont, its certificates are supersets to the ones of \ucont.
This follows from \cwcont\ reducing to \ucont by taking $\mathcal{S} {\supseteq} \cup \CalS_i$, \ie, overapproximating some of the classes with a larger $\mathcal{S}$.
This is analogous to setting $L{\geq}\max L_i$ in the Lipschitz case.
Then, from \cref{thm:polar_set_properties}\textit{iv}, it directly follows that \cwcont\ certificates are always supersets of \ucont\ certificates.
Another view is that \ucont\ certificates are restricted to only symmetric sets since $\CalS{\oplus}\sm\CalS$ is symmetric (\cref{thm:always_symmetric}), while
\cwcont  certificates, \ie, $\bigcap_{i\neq c_A} ( \CalS_i {\oplus}\sm\CalS_{c_A})^{r_i}$, can be asymmetric.%

The example in \cref{fig:2d_linear_classifier} (with detailed calculations in \cref{sec:2d_linear_classifier_workedout}) shows how the certified regions can vary depending on whether we use \SLip or Lipschitz certificates and on the \cwcont or \ucont modes.

\begin{theoremEnd}[all end]{auxlemma}
    \label{thm:always_symmetric}
    For any bounded set $\CalS\subset\RR^d$ it holds that $\CalS\oplus-\CalS$ is symmetric, \ie
    $$x\in(\CalS\oplus-\CalS) \implies -x\in(\CalS\oplus-\CalS).$$
    Furthermore, for any $r>0$ and any symmetric $\CalS\subset\RR^d$, it holds that $\CalS^r$ is also symmetric.
    Finally, if $\CalS$ is symmetric and convex, then
    $$\CalS\oplus-\CalS=2\CalS.$$
\end{theoremEnd}
\begin{proofEnd}
    If $x\in(\CalS\oplus-\CalS)$ then $\exists s_1,s_2\in\CalS$ such that $x=s_1-s_2$.
    However, then it also must hold that $s_2-s_1=-x$ is in $\CalS\oplus-\CalS$.
    
    Let's now prove the second part.
    The condition for $-y$ to be in $\CalS^r$ when $\CalS$ is symmetric is $\sup_{x\in\CalS} x^\top (-y) \leq r$.
    The left side can be rewritten as:
    $$
        \sup_{x\in\CalS} x^\top (-y) =
        \sup_{x\in\CalS} (-x)^\top y =
        \sup_{x\in-\CalS} x^\top y =
        \sup_{x\in\CalS} x^\top y,
    $$
    which is the same as the condition for $y$ to be in $\CalS^r$.
    In the above, we use the fact that $\CalS=-\CalS$, the definition of $\CalS$ being symmetric.   

    For the last part we have
    $$\CalS\oplus-\CalS=S\oplus S = 2\CalS.$$
    The last equality follows from convexity: for any $s_1, s_2\in \CalS$ it holds that $(s_1+s_2)/2\in\CalS$ and hence $s_1+s_2\in 2\CalS$.
\end{proofEnd}

\begin{theoremEnd}[all end]{auxlemma}
    \label{thm:polar_as_hyperplane_intersection}
    For a $\CalS\subset\RR^d$ its polar set of radius $r$ is the intersection of $|\CalS|$ half-spaces:
    $$(\CalS)^r = \bigcap_{s\in\CalS} \left\{ x\in\RR^d :  \frac{1}{r} s^\top x \leq 1 \right\}.$$
\end{theoremEnd}

\subsection{Tightening Certificates via Class Differences}

\label{sec:class_difference}

We conclude this section by showing how to further enlarge the certificates by directly targeting the \CalS-Lipschitzness of the class difference.
Recall the \Scert $Q=\bigcap_{i\neq c_A} (\CalS_i \oplus - \CalS_{c_A} )^{r_i}$ for the \cwcont mode from \cref{thm:Slip_certificate}.
The role of the $\CalS_i \oplus - \CalS_{c_A}$ term is to measure the \SLip continuity of $h_{i\sm c_A}=f_i-f_{c_A}$.
It is straightforward to see that $h_{i\sm c_A}$ is indeed $(\CalS_i \oplus - \CalS_{c_A})$-Lipschitz. 
However, it is not necessarily the tightest $\CalS$ for $h_{i\sm c_A}$.
Intuitively, $\CalS_i \oplus - \CalS_{c_A}$ takes the differences of the gradients of $f_i$ and $f_{c_A}$, \emph{regardless} of the input $x$.
However, the set of gradients of $h_{i\sm c_A}$ are the difference of gradients of $f_i$ and $f_{c_A}$ \emph{at the same} $x$.
If all classes are similarly sensitive at a given $x$ but their sensitivity varies \emph{jointly} across the domain, the difference between $\CalS_i \oplus - \CalS_{c_A}$  and the gradients of $h_{i\sm c_A}$ can be significant.
Using this, we can tighten \cref{thm:Slip_certificate} with class-difference (\cdcont) certificates.
\begin{theoremEnd}[restate, end]{theorem}
    \label{thm:Slip_certificate_diff}
    Let $f:\RR^d\to\RR^K$ be a classifier such that $h_{i\sm j} = f_i-f_j$ is $\CalS_{i\sm j}$-Lipschitz, $\forall i,j\in 1,\ldots,K,  i\neq j$.
    Then, given an input $x\in\RR^d$, $f$ is robust at $x$ against all $\delta$ in $Q = \bigcap_{i\neq c_A} (\CalS_{i\sm c_A})^{r_i}$.
\end{theoremEnd}
\begin{proofEnd}
    For a fixed class $i\neq c_A$ we have that the following must hold from \cref{def:Slip}:
    \begin{align*}
        h_{i\sm c_A} (y) - h_{i\sm c_A} (x) &\leq \rho_{\CalS_{i\sm j}} (y-x) \\
        f_i(y)-f_{c_A}(y)-f_i(x)+f_{c_A}(x) &\leq \rho_{\CalS_{i\sm j}} (y-x).
    \end{align*}
    Rearranging the terms gives:
    $$
        f_i(y)-f_{c_A}(y) \leq \rho_{\CalS_{i\sm j}} (y-x) + \underbrace{f_i(x)-f_{c_A}(x)}_{-r_i}.
    $$
    We are interested in the values of $y$ for which the left-hand side is nonpositive as these are inputs for which the confidence is higher for $c_A$ than for $i$.
    Hence, we restrict the right-hand side to be upper-bounded by zero:
    $$
        \rho_{\CalS_{i\sm j}} (\underbrace{y-x}_{\delta}) \leq r_i.
    $$
    The values of $\delta$ satisfying this inequality are exactly the polar set $(\CalS_{i\sm j})^{r_i}$ (\cref{def:polar_set}).
    
    Finally, as we need that the confidence for class $c_A$ is larger than the confidences for any other class, we need to take the intersection over $i\neq c_A$ resulting in the certificate $Q = \bigcap_{i\neq c_A} (\CalS_{i\sm c_A})^{r_i}$.
\end{proofEnd}
The following \Cref{ex:class_difference} illustrates how the \cdcont\ certificates  (\cref{thm:Slip_certificate_diff}) are larger than the \cwcont\ certificates (\cref{thm:Slip_certificate}).
\begin{example}
    \label{ex:class_difference}
    Consider the piece-wise linear classifier $f:\mathbb{R} \rightarrow \mathbb{R}^2$ that we wish to certify at $x_0=2$:
    \begin{minipage}{0.48\columnwidth}
        \begin{align*}
            f_1(x){=}%
            \begin{cases}
                0.1x {+} 0.7 & \text{if } x{\leq}3, \\
                1.1x {-} 2.3 & \text{if } x{>}3,
            \end{cases}\\
            f_2(x){=}%
            \begin{cases}
                0.3x {+} 0.1 & \text{if }x {\leq}3, \\
                1.3x {-} 2.9 & \text{if }x {>} 3.
            \end{cases}
        \end{align*}
    \end{minipage}%
    \begin{minipage}{0.48\columnwidth}
        \includegraphics[width=\textwidth]{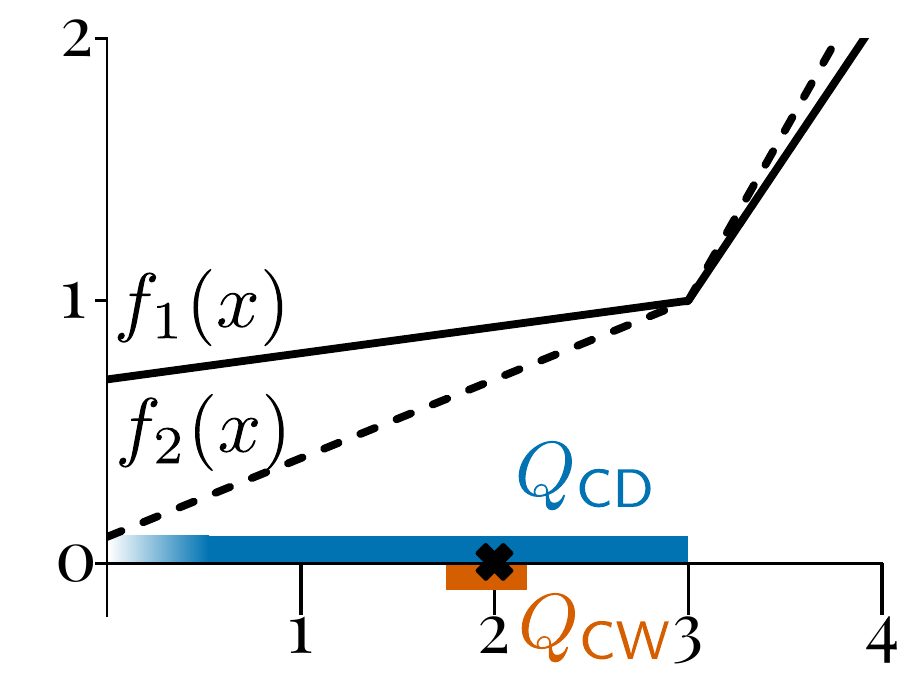}
    \end{minipage}
    We have $c_A=1$, $r_2=0.2$, $\CalS_1=\{0.1, 1.1\}$, $\CalS_2=\{0.3, 1.3\}$, $\CalS_2 \oplus-\CalS_1=\{0.2,-0.8,1.2,0.2\}$, $\CalS_{2-1}=\{0.2\}$.
    Therefore, \cref{thm:Slip_certificate} gives a certificate $Q_\textsf{CW}=(\CalS_2 \oplus-\CalS_1)^{r_2}=[\nicefrac{0.2}{-0.8}, \nicefrac{0.2}{1.2}]$.
    \cref{thm:Slip_certificate_diff} instead gives the much bigger $Q_\textsf{CD}=(\CalS_{2\sm 1})^{r_2}=(-\infty, 1]$.
\end{example}

\begin{figure}
    \centering
    \resizebox{1.04\columnwidth}{!}{
        \begin{tikzpicture}[node distance=10cm,line width=1pt]
            \node(SCD)    at (0,3) [text width=3.5cm,align=center] {Class-difference (\cdcont)\\\SLip\\(\cref{thm:Slip_certificate_diff})};
            \node(LCD)    at (0,0) [text width=3.5cm,align=center] {Class-difference (\cdcont)\\Lipschitz\\\citep{weng2018evaluating,yang2022on}};

            \node(SCW)    at (5.5,3) [text width=3cm,align=center] {Class-wise (\cwcont)\\\SLip\\(\cref{thm:Slip_certificate})};
            \node(LCW)   at (5.5,0) [text width=3cm,align=center] {Class-wise (\cwcont)\\Lipschitz\\(\cref{thm:bounded_gradients_certificates})};

            \node(SU)   at (11,3) [text width=3cm,align=center] {Uniform (\ucont)\\\SLip\\(\cref{thm:Slip_certificate})};
            \node(LU)  at (11,0)  [text width=3cm,align=center] {Uniform (\ucont)\\ Lipschitz\\(\cref{thm:bounded_gradients_certificates})};

            \draw[->](SCD) -- (LCD)  node[midway, left,align=center, font = {\small\itshape}] {Take $L_{i\sm j}$ to be\\ $\sup\norm{\CalS_{i\sm j}}_\star$}  ;
            \draw[->](SCW) -- (LCW)  node[midway, left,align=center, font = {\small\itshape}] {Take $L_{i}$ to be\\ $\sup\norm{\CalS_{i}}_\star$}  ;
            \draw[->](SU) -- (LU)  node[midway, left,align=center, font = {\small\itshape}] {Take $L$ to be\\ $\sup\norm{\CalS}_\star$}  ;
            
            \draw[->](SCD) -- (SCW)  node[midway, above,align=center, font = {\small\itshape}] {Replace $\CalS_{i\sm j}$\\ with $\CalS_i{\oplus}{\sm}\CalS_j$}  ;
            \draw[->](SCW) -- (SU)  node[midway, above,align=center, font = {\small\itshape}] {Replace $\CalS_i$\\ with $\cup \CalS_i$}  ;
            
            \draw[->](LCD) -- (LCW)  node[midway, above,align=center, font = {\small\itshape}] {Replace $L_{i\sm j}$\\ with $L_i+L_j$}  ;
            \draw[->](LCW) -- (LU)  node[midway, above,align=center, font = {\small\itshape}] { Replace $L_i$\\ with $\max L_i$}  ;
        \end{tikzpicture}
    }

    \caption{The lattice of continuity certificates. $A \to B$ means that the certificate provided by $B$ is a subset of the certificate of $A$. Therefore, class-difference \Scerts are the largest, while uniform Lipschitz certificates are the smallest.}
    \label{fig:smoothness_lattice}
\end{figure}
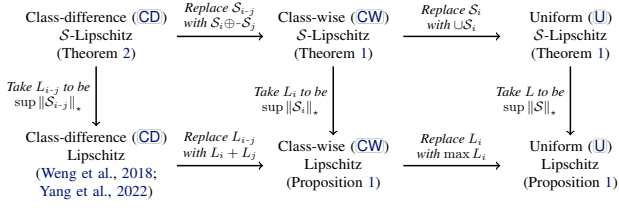

This approach generalizes the \cwcont\ \Scerts from \cref{thm:Slip_certificate} and provides the tightest certificates.
For example, replacing $\CalS_{i\sm c_A}$ with $(\CalS_i {\oplus}{\sm}\CalS_{c_A})$ recovers \cref{eq:Slip_certificate_cw}.
Hence, throughout the rest of the paper, we will use class difference unless stated otherwise.
Prior work looked at the Lipschitz \cdcont\ certificates \citep{weng2018evaluating} and regularization \citep{yang2022on}.
To the best of our knowledge, we are the first to offer a theoretical justification of why it enlarges the certificates through the new lens of \SLipness.

\cref{fig:smoothness_lattice} summarizes the big picture relating the certificates with function continuity and positions our new results with respect to prior art.
Our results fully complete the lattice relating all components together, \ie, Lipschitz, \SLip, \cwcont, \ucont, and \cdcont\ modes, and their relation to certification.
The bottom row shows the Lipschitz certificates, while the top row shows our \Scerts. %
The vertical arrows demonstrate how \Scerts are always larger than the corresponding Lipschitz certificates.
The horizontal arrows show that \cwcont certificates are smaller than \cdcont\ certificates, %
and that \ucont certificates are smaller than \cwcont certificates.
Therefore, the \cdcont\ \Scerts we introduce here provide the largest certificates (top left corner), while the \ucont Lipschitz certificates (bottom right) ---which are commonly used in prior work--- result in the smallest certificates. 

\section{Robustness of Ensembles of Classifiers}
\label{sec:smooth_ensembles}

We can use \SLipness to study how the robustness properties of individual classifiers affect the robustness of an ensemble of them.
Given $N$ classifiers $f^j:\RR^d\to\RR^K$, consider their weighted ensemble:
\begin{equation}
    g(x) = \textstyle{\sum_{j=1}^N} \alpha_j f^j (x), ~~ \alpha_j \geq 0,~ \sum_{j=1}^N \alpha_j =1.%
    \label{eq:ensemble_definition}
\end{equation}
We will indicate the prediction gaps of $f^j$ as $r^j$.
We can use the \Scerts from \cref{thm:Slip_certificate_diff} %
in order to relate the ensemble robustness to that of the individual classifiers.
\begin{theoremEnd}[all end]{auxlemma}[Scaling of \SLip Classifiers]
    \label{thm:slip_scaling}
    Consider a constant $\alpha>0$ and a classifier $f:\RR^d\to\RR^K$ such that $h_{i\sm j}=f_i-f_j$ is $\CalS_{i\sm j}$-Lipschitz for all $i\neq j$.
    Then $\alpha h_{i\sm j}=\alpha f_i-\alpha f_j$ is $\alpha \CalS_{i\sm j}$-Lipschitz but $f$ and $\alpha f$ have the same certificates:
    \begin{equation}
        Q_f = \bigcap_{i\neq c_A^g} ( \CalS_{i\sm c_A^g})^{r_i} = \bigcap_{i\neq c_A^g} ( \alpha \CalS_{i\sm c_A^g})^{\alpha r_i} = Q_{\alpha f}.
        \label{eq:scaling_classifier_classwise}
    \end{equation}
\end{theoremEnd}
\begin{proofEnd}
    Note that scaling with a positive constant $\alpha$ does not change the top class $c_A^g$ and also scales the prediction gaps proportionally: $\alpha f_{c_A^g} (x) - \alpha f_i(x) = \alpha r_i$.
    
    Next, let's show that $\alpha f_i$ is $\alpha\CalS_i$-Lipschitz.
    If $g:\RR^d\to\RR$ is \SLip, then we have
    $$ - \sup_{c\in\CalS} c^\top (x-y) \leq g(y)-g(x) \leq \sup_{c\in\CalS} c^\top (y-x)$$
    for all $x,y\in\RR^d$.
    As $\alpha > 0$, multiplying everything by $\alpha$ results in:
    \begin{gather*}
        - \sup_{c\in\CalS} \alpha c^\top (x-y) \leq \alpha g(y)-\alpha g(x) \leq \sup_{c\in\CalS} \alpha c^\top (y-x) \\
         - \sup_{c\in\alpha\CalS} c^\top (x-y) \leq \alpha g(y)-\alpha g(x) \leq \sup_{c\in\alpha\CalS} c^\top (y-x),
    \end{gather*}
    which is the condition for $\alpha g$ being $(\alpha\CalS)$-Lipschitz.

    Now, we can show that scaling the $\CalS$ set and the polar set radius with the same constant does not change the polar set:
    \begin{align*}
        (\alpha\CalS)^{\alpha r}
        &= \left\{ y\in\RR^d ~:~ \sup_{x\in\alpha\CalS}\ x^\top y \leq \alpha r \right\} \\
        &= \left\{ y\in\RR^d ~:~ \sup_{x\in\CalS}\ \alpha x^\top y \leq \alpha r \right\} \\
        &= \left\{ y\in\RR^d ~:~ \sup_{x\in\CalS}\ x^\top y \leq r \right\} \\
        &= (\CalS)^r.
    \end{align*}
    \cref{eq:scaling_classifier_classwise} directly follows. 
\end{proofEnd}
\begin{theoremEnd}[end, restate]{theorem}[Addition of \SLip classifiers]
    \label{thm:slip_addition}
    Take an ensemble as in \cref{eq:ensemble_definition} with $N=2$ and the  \cdcont\ setting, \ie,
    $h^j_{i\sm k}{=}f^j_i-f^j_k$ is $\CalS_{i\sm k}^{j}$-Lipschitz.
    Then, at a fixed $x\in\RR^d$, it holds that $g$ is robust against all $\delta$ in
    $$ Q_g = \textstyle{\bigcap_{i\neq c_A^g}} \left( \alpha_1 \CalS_{i\sm c_A^g}^{1} \oplus \alpha_2 \CalS_{i\sm c_A^g}^{2} \right)^{r_i^g},$$
    with $c_A^g=\arg\max_{i} g_i$ and $r_i^g=g_{c_A^g} - g_i $.
    The case for $N>2$ follows by induction.
\end{theoremEnd}
\begin{proofEnd}
    From \cref{thm:slip_scaling} we know that $\alpha_1 h_1^{i\sm j}$ and $\alpha_2 h_2^{i\sm j}$ are $\alpha_1\CalS_{i\sm j}^{1}$- and $\alpha_2\CalS_{i\sm j}^{2}$-Lipschitz.
    Then, following \cref{thm:gradients_to_S}, we have that $\nabla \alpha_1 h_1^{i\sm j} \in \hull\alpha_1\CalS_{i\sm j}^{1}$ and $ \nabla \alpha_2 h_2^{i\sm j} \in \hull\alpha_2\CalS_{i\sm j}^{2}$.
    Since $\nabla h^g_{i\sm j} = \nabla ( g_i-g_j ) = \nabla \alpha_1 h_1^{i\sm j} + \nabla \alpha_2 h_2^{i\sm j}$, we have $\nabla h^g_{i\sm j} \in \alpha_1\hull\CalS_{i\sm j}^{1} \oplus \alpha_2\hull\CalS_{i\sm j}^{2} = \hull(\alpha_1\CalS_{i\sm j}^{1}\oplus\alpha_2\CalS_{i\sm j}^{2})$ as constructing the convex hull and taking the Minkowski sum commute.
    By \Cref{thm:Slip_certificate}, $h^g_{i\sm j}$ is $(\hull(\alpha_1\CalS_{i\sm j}^{1}\oplus\alpha_2\CalS_{i\sm j}^{2}))$-Lipschitz which by \Cref{thm:hull_can_be_dropped} is the same as being $(\alpha_1\CalS_{i\sm j}^{1}\oplus\alpha_2\CalS_{i\sm j}^{2})$-Lipschitz.
    The rest follows from \Cref{thm:Slip_certificate_diff}.
\end{proofEnd}
In the \ucont\ mode, where all classes have the same Lipschitzness $\mathcal{S}^{j} \supseteq \cup_i \mathcal{S}_i^{j}$ the $\mathcal{S}_{i\sm k}^{j}$ term reduces to $\mathcal{S}^{j} \oplus -\mathcal{S}^{j}$.

We study whether ensembling two classifiers $f_1$ and $f_2$ results in better robustness by comparing the ensemble certificate $Q_g$ with the individual certificates $Q_1$ and $Q_2$. %
We identify three regimes. 
First, the ensemble certificate $Q_g$ includes all certified points in $Q_1$ and $Q_2$. 
Second, the ensemble certificate fails to include some perturbations certified in both $Q_1$ and $Q_2$. 
Third, an ensemble certificate somewhere between the two. Formally,
\begin{align}\refstepcounter{equation}
    &\ Q_g \supset Q_1 \cup Q_2 &&\text{uniform improvement,} \tag*{\Ronesymb} \label{R1}\\
    Q_1 \cap Q_2 \subseteq &\ Q_g \subseteq Q_1 \cup Q_2 &&\text{inconclusive,} \tag*{\Rtwosymb} \label{R2} \\
    Q_1 \cap Q_2 \supset &\ Q_g  &&\text{uniform reduction}. \tag*{\Rthreesymb} \label{R3}
\end{align}
Ideally, we wish to construct ensembles that are in regime \Rone. 
We may tolerate ensembles in \Rtwo. 
But most importantly, we want to avoid ensembles in regime \Rthree\ at all costs.

The certification regime depends on whether we are in the \ucont\ or \cdcont\ mode.
It also depends on the ensemble agreement on the top predictions, \ie, which of the following holds:
\begin{align}\refstepcounter{equation}
    c_A &= c_A^j = \arg\max_{i} f^j_i(x) , &&\text{for all } j\in1,\dots,N \tag*{\samecasymb} \label{samecatarget} \\
    c_A^j &\neq c_A^{j'}, &&\text{for } j\neq j' \tag*{\diffcasymb} \label{diffcatarget} \\
    c_B &= c_B^j = \arg\max_{i\neq c_A^j} f^j_i(x) , &&\text{for all } j\in1,\dots,N \tag*{\samecbsymb} \label{samecbtarget} 
\end{align}
\refstepcounter{equation}%
The rest of this section outlines the conditions leading to each one of the \Rone,\Rtwo\ and \Rthree\ certification regimes.

Let us first examine a common scenario for ensembles and identify what certification regime most ensembles fall in. In particular, consider the setting where the constituent classifiers agree on the top two predictions (\sameca and \samecb). 
This is a reasonable assumption, particularly when the number of constituent classifiers $N$ is small and the training procedure for all classifiers is similar. %
Under the common \ucont mode where all classes are similarly Lipschitz, one might guess that ensembling such agreeing classifiers must boost robustness.
However, the above conditions put the ensemble solidly in regime \Rtwo, as shown in \cref{thm:same_ca_cb}.

\textEnd{\begin{figure*}
    \begin{equation}
        \label{eq:lincom_intersection_union_Qg}
        Q_g = \left\{ 
        y\in\RR^n :
        \sup_{\substack{x_1,x_2\in\CalS^{1}\\x_3,x_4\in\CalS^{2}}} 
            \left\{ \alpha_1(x_1-x_2)^\top y + \alpha_2(x_3-x_4)^\top y \right\}
            \leq \alpha_1 r^1 + \alpha_2 r^2
        \right\}
    \end{equation}
    \end{figure*}}
\begin{theoremEnd}[end, restate]{theorem}
    \label{thm:same_ca_cb}
    Consider an ensemble of \ucont\ classifiers and a fixed $x$ for which \sameca\ and \samecb\ hold. 
    Then, for any choice of weights $\alpha_j$ in \cref{eq:ensemble_definition}, the \Scert of the ensemble satisfies \Rtwo.
\end{theoremEnd}
\begin{proofEnd}
    We will prove only the case for $N=2$.
    $N>2$ follows by induction.
    Furthermore, we assume $\alpha_j\geq 0, \forall j$ as in \cref{eq:ensemble_definition}.
    
    We will denote the individual classifier gaps and the ensemble gap as $r^1=f^1_{c_A}(x)-f^1_{c_B}(x)$, $r^2=f^2_{c_A}(x)-f^2_{c_B}(x)$, $r^g=g_{c_A}(x)-g_{c_B}(x)$.
    First, from \cref{thm:Slip_certificate} we have
    \begin{align*}
        Q_1 &= (\CalS^{1}\oplus-\CalS^{1})^{r^{1}}, \\
        Q_2 &= (\CalS^{2}\oplus-\CalS^{2})^{r^2}, \\
        Q_g &= \left( (\alpha_1 \CalS^{1} \oplus \alpha_2 \CalS^{2} ) \oplus - (\alpha_1 \CalS^{1} \oplus \alpha_2\CalS^{2} ) \right)^{\alpha_1 r^1 + \alpha_2 r^2} \\
            &= (\alpha_1 (\CalS^{1}\oplus -\CalS^{1}) \oplus \alpha_2 (\CalS^{2}\oplus-\CalS^{2}))^{\alpha_1 r^1 + \alpha_2 r^2}.
    \end{align*}
    $Q_g$ can also be expanded as \cref{eq:lincom_intersection_union_Qg}.
    Consider the two inequalities that define $Q_1$ and $Q_2$:
    $$ \sup_{x_1,x_2\in\CalS^{1}} (x_1-x_2)^\top y \leq r^1, \hfill \sup_{x_3,x_4\in\CalS^{2}} (x_3-x_4)^\top y \leq r^2.$$
    If for a given $y$, both of these hold, then the inequality in \Cref{eq:lincom_intersection_union_Qg} also must hold.
    Hence, the intersection of $Q_1$ and $Q_2$ must be a subset of $Q_g$.
    Similarly, it is necessary for at least one of them to hold, hence every element of $Q_g$ must be an element of the union of $Q_1$ and $Q_2$.
\end{proofEnd}

\begin{theoremEnd}[all end]{auxlemma}
    \label{thm:sum_of_norm_balls}
    Let $S\subseteq\RR^d$ be a convex set, $\alpha, \beta \geq 0$, and $a,b\in\RR^d$.
    Then it holds that
    $$(\alpha S + a) \oplus (\beta S + b) = (\alpha+\beta)S + (a+b).$$
    A special case is the summing of $\ell_p$ norm balls ($p\geq 1$):
    $$\mathcal B_p[\mu_1, \epsilon_1] \oplus \mathcal B_p[\mu_1, \epsilon_1] = \mathcal B_p[\mu_1+\mu_2, \epsilon_1+\epsilon_2].$$
\end{theoremEnd}
\begin{proofEnd}
    It is trivial to see that
    $$(\alpha S + a) \oplus (\beta S + b) = \alpha S \oplus \beta S  + (a+b).$$
    Hence, we only need to show if $\alpha S \oplus \beta S \stackrel{?}{=} (\alpha+\beta)S$ which is the same as:
    $$ S_L=\left\{ \alpha x + \beta y : x,y\in S \right\} \stackrel{?}{=} \left\{ (\alpha + \beta) x' : x'\in S \right\}=S_R.$$
    
    It is obvious that $S_R\subseteq S_L$.
    So we only need to show that $S_L\subseteq S_R$.
    Take a $z\in S_L$. Then there must be $x,y\in S$ such that $\alpha x + \beta y = z$.
    Now, take $x'=\nicefrac{z}{(\alpha+\beta)}$:
    $$x' = \frac{z}{\alpha+\beta} = \frac{\alpha x + \beta y}{\alpha+\beta} = \frac{\alpha}{\alpha+\beta}x + \frac{\beta}{\alpha+\beta}y.$$
    $x'$ is a linear combination of elements of the convex $S$, hence $x'$ is also in $S$.
    Therefore, for every $z\in S_L$ we can construct an $x'\in S$ such that $(\alpha+\beta)x'=z \in S_R$.
    This concludes our proof that $S_L=S_R$.

    The $\ell_p$ norm ball special case follows directly when we note that $\mathcal B_p[\mu, \epsilon]$ can be represented as
    $$ \epsilon \cdot \{ x\in\RR^d : \|x\|_p \leq 1 \} + \mu. $$
\end{proofEnd}
\begin{theoremEnd}[all end]{auxlemma}
    \label{thm:polar_norm_balls}
    Take to be $\mathcal B_\star\subset\RR^d$ a closed convex symmetric set.
    Define $\mathcal B$ to be the norm ball of its dual norm, \ie:
    $$\mathcal B = \left\{ y\in\RR^d : \sup_{x\in \mathcal B_\star} x^\top y \leq 1\right\}. $$
    Then, the polar set of $\epsilon \mathcal B$ with radius $r$ is:
    $$ (\epsilon \mathcal B_\star)^r = \frac{r}{\epsilon} \mathcal B. $$
\end{theoremEnd}

\Cref{thm:same_ca_cb} is particularly concerning when $\CalS^{1}$ and $\CalS^{2}$ are norm balls with the same norm but different radii, as we show with an example in \cref{ex:same_ca_cb_same_shape}.

Under the assumptions in \cref{thm:same_ca_cb} ensembling can never be in the favourable regime \Rone.
The following section shows how relaxing these conditions enables all three regimes. 

\subsection{Certification Governed by the Prediction Gap}
\label{sec:ensemble_same_S}

\begin{figure*}
    \centering
    \includegraphics[width=0.9\textwidth]{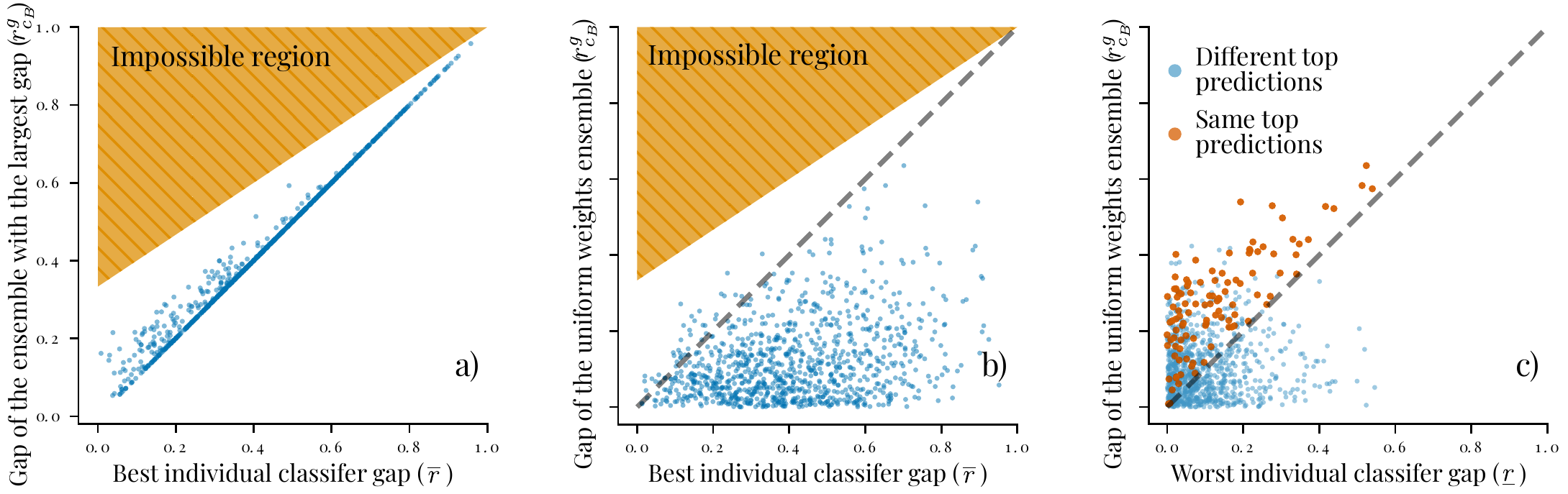}
    \caption{
    A set of 1000 ensembles of 2, 3 or 4 classifiers, each a uniform draw from the 4-dimensional probability simplex.
    (\textbf{a}) shows the best individual gap among the classifiers in each ensemble ($\overline{r}$) vs the largest ensemble gap ($r^g_{c_B}$) attainable across all $\alpha_j$. 
    The larger the best gap $\overline r$, the lower the potential gain $r^g_{c_B}-\overline{r}$ (the vertical gap between the diagonal and the impossible region).
    (\textbf{b}) has the same horizontal axis as a) but the ensemble gap ($r^g_{c_B}$) is computed for uniform weights $\alpha_j$.
    Most of the uniform weights ensembles witness gain loss.
    (\textbf{c}) has the same vertical axis as b) but the horizontal axis shows the worst individual gap ($\underline{r}$) instead of the best one. 
    The ensembles with same (\sameca) and different top predictions (\diffca) are highlighted, showing that the \sameca\ regime always results in $r^g_{c_B} \geq \underline{r}$.
}
    \label{fig:gaps_illustrations}
\end{figure*}
\begin{figure}
    \includegraphics[width=\columnwidth]{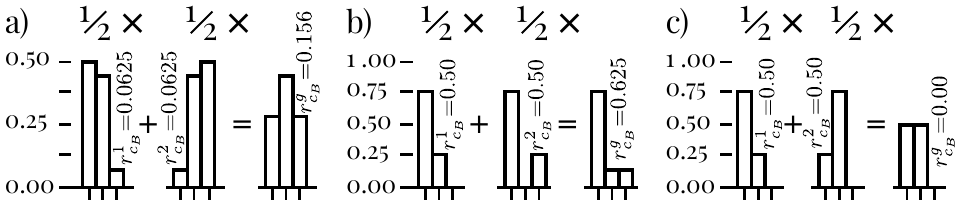}
    \vspace{-0.50cm}
    \caption{Ensembles of two classifiers ($N{=}2, K{=}3$) in regime \rone\ (\textbf{a} and \textbf{b}), and regime \rthree\ with $r^g=0$ and hence $Q_g=\{0\}$ (\textbf{c}).}
    \label{fig:regime_1_example}
    \vspace{-0.25cm}
\end{figure}

\cref{thm:slip_addition} shows that the prediction gaps $r$ and the continuity $\CalS$ interact in complex ways in the construction of the ensemble certificate $Q_g$.
However, if all classifiers have the same smoothness for all the classes, i.e.\ \ucont\ and $\CalS^{j} = \CalS$, then the differences between $Q_1$, $Q_2$ and $Q_g$ are fully determined by $r^1$, $r^2$ and $r^g$.
We will refer to this setting as \ucontstar.
This restriction is not uncommon as often ensembled classifiers are identically trained.
For example, if randomized smoothing is used, then $\CalS$ is uniquely defined by the smoothing distribution \citep{yang20randomized,eiras2022ancer,rumezhak2023rancer}, which is the same for all constituents.

In this case, there is one-to-one mapping between the certification regimes \Rone, \Rtwo, \Rthree\, and the prediction gaps. 
Consider the following conditions on the prediction gaps:
\begin{align}\refstepcounter{equation}
    &\ r^g_{c_B} > {\textstyle \max_{j} r^{j}_{c_B}} = \overline{r}  &&\text{gap gain,}  \label{r1}\tag*{\ronesymb}\\
    \underline{r} \leq &\ r^g_{c_B} \leq \overline{r} &&\text{inconclusive,} \label{r2}\tag*{\rtwosymb}  \\
    {\textstyle \min_{j} r^{j}_{c_B}} = \underline{r} > &\ r^g_{c_B}  &&\text{gap loss}. \label{r3}
\tag*{\rthreesymb} 
\end{align}
Then, we have that \ronesymb $\implies$ \Rone, \rtwosymb $\implies$ \Rtwo, and \rthreesymb $\implies$ \Rthree. 
Therefore, in this subsection, we will focus on the conditions resulting in \ronesymb, \rtwosymb, and \rthreesymb towards understanding the certification properties of the ensembles in mode \ucontstar.

\stress{Similar top two predictions results in \rtwo.}
Note that if the top predictions are consistent across all constitute classifiers, \ie \sameca\ and \samecb\ hold, this implies that the ensemble prediction gap is the linear combination of the individual predictions gaps  $r^g_{c_B}=\sum_j \alpha_j r^j_{c_B}$. 
Hence, the gap regime must be \rtwo\ as $\min_j r_{c_B}^j \leq r^g_{c_B} \leq \max_j r_{c_B}^j$, which implies regime \Rtwo\ for \ucontstar.
This is a special case of \Cref{thm:same_ca_cb}.

\stress{Regime \Rone\ is possible.}
For a \ucontstar\ ensemble, prediction gaps in regime \rone\ ($r^g_{c_B} {>} \overline r$) imply \Rone. %
One conditions for \Rone is \diffca\ and \samecb\ with the classifiers having similar confidences in the top two classes and low confidence in all other classes (see \cref{fig:regime_1_example}a). 
Another possibility is \sameca, but each classifier having a different second prediction, as in \cref{fig:regime_1_example}b.

\stress{The margin of improvement when \rone\ holds is small.}
Although the feasibility of regime \rone\ is noteworthy, unfortunately, the improvement of $r^g_{c_B}$ over $\overline r$ is limited. %

\begin{theoremEnd}[restate, end]{proposition}
    \label{thm:gap_gain_upper_bound}
    Consider $N$ classifiers over $K$ classes. We have that for any ensemble $g$ the prediction confidence is upper bounded as follows:
    \begin{equation}
        r^g_{c_B} \leq \overline r + \frac{1-\overline r}{2} - \frac{1-\overline r}{2(K-1)}
        \label{eq:gap_gain_upper_bound}
    \end{equation}
    The bound is tight: given $\overline{r}$ and $K$ there exists an ensemble $f_1,\ldots,f_N$, such that the prediction gap $r^g_{c_B}$ of $g$ attains the upper bound.
\end{theoremEnd}
\begin{proofEnd}
    We have $N$ normalized $K$-class classifiers, so $f_j\in\RR^K$, $\sum_i f^j_i = 1$, $f^j_i >0$, for all $j\in 1,\ldots,N$.
    Furthermore, $g$ being a linear combination of $f^1,\ldots,f^N$ means that we have $g=\sum_{j=1}^N \alpha_j f_j$, for some $\alpha_j{>}0, \sum_j \alpha_j = 1$.
    $c_A^j$ and $c_B^j$ are the top two classes of $f_j$ and similarly $c_A^g$ and $c_B^g$ are the top two classes of $g$.
    We also have $r^j=f^j_{c_A^j}-f^j_{c_B^j}$, $r^g=g_{c_A^g}-g_{c_B^g}$, and $\overline r = \max_{j\in 1,\ldots,N} r^j$.

    From $f^j_{c_A^j} + f^j_{c_B^j} \leq 1$, $f^j_{c_A^j} \geq f^j_{c_N^j}$ and $f^j_{c_A^j}-f^j_{c_B^j} \leq \overline r$ we have:
    \begin{equation*}
        f^j_{c_A^j} \leq \overline r + \frac{1-\overline r}{2}.
    \end{equation*}
    Note that $g_{c_A^g}$ has the same bound:
    \begin{equation}
        \begin{aligned}
             g_{c_A^g} &= \sum_j \alpha_j f^j_{c_A^g} \\
             &\leq \sum_j \alpha_j f^j_{c_A^j} \\
             &\leq \sum_j \alpha_j \left(\overline r + \frac{1-\overline r}{2} \right) \\
             &= \overline r + \frac{1-\overline r}{2}.
        \end{aligned}
        \label{eq:gap_gain_upper_bound_cA_bound}
    \end{equation}
   
    As the classes must sum to 1 we have
    $$\sum_{i\neq c_A^j} f^j_i = 1 - f^j_{c_A^j} \geq \frac{1-\overline r}{2}, ~\forall j\in 1,\ldots,N. $$
    The minimum $c_B^g$ can be obtained when all $c_A^j$ are the same.
    Then the top weight for each classifier gets sent to $c_A^g$.
    Therefore, we have:
    \begin{equation*}
        \begin{aligned}
            \sum_{i\neq c_A^g} g_i &= \sum_{i\neq c_A^g} \sum_j \alpha_j f^j_i \\
            &= \sum_j \alpha_j \sum_{i\neq c_A^j} f^j_i \\
            &\geq \sum_j \alpha_j \frac{1-\overline r}{2} \\
            &= \frac{1-\overline r}{2}.
        \end{aligned}
    \end{equation*}
    The largest element of $\{ g_i : i\neq c_A^g \}$ must be at least as large as the average, hence:
    \begin{equation}
        g_{c_B^g} \geq \frac{1-\overline r}{2(K-1)}
        \label{eq:gap_gain_upper_bound_cB_bound}
    \end{equation}
    
    Hence, from \cref{eq:gap_gain_upper_bound_cA_bound,eq:gap_gain_upper_bound_cB_bound} we have:
    $$r^g = g_{c_A^g}-g_{c_B^g} \leq \overline r + \frac{1-\overline r}{2} - \frac{1-\overline r}{2(K-1)}.$$

    Let's show that this bound is tight.
    For that, we need to construct a set of $N$ classifiers that attain it.
    Consider $N=K-1$: the number of classifiers being one less than the number of classes. 
    Take all $f_j$ to be such that 
    $$
    f^j_i=
    \begin{cases}
        \bar{r} + (1-\bar{r})/2 &\text{if } i=K, \\
        (1-\bar{r})/2     &\text{if } i\leq K-1, i=j, \\
        0           &\text{if } i\leq K-1, i\neq j.
    \end{cases}
    $$
    It is easy to verify that $\sum_{i\in 1,\ldots,K} f^j_i = 1, ~\forall j\in 1,\ldots,N$.
    Take also uniforms weights: $\alpha_j = \nicefrac{1}{N}$.
    Then we have:
    $$
    g_i = 
    \begin{cases}
        \frac{1-\bar{r}}{2N}=\frac{1-\bar{r}}{2(K-1)} &\text{if } i\leq K-1, \\
        \bar{r}+\frac{1-\bar{r}}{2} &\text{if } i=K.
    \end{cases}
    $$
    And hence: $r^g = \bar{r}+\frac{1-\bar{r}}{2} - \frac{1-\bar{r}}{2(K-1)}$.
\end{proofEnd}

\Cref{eq:gap_gain_upper_bound} does not depend on the weights $\alpha_j$.
Furthermore, $r_{c_B}^g - \overline{r}$ decreases monotonically with $\overline{r}$, reaching $0$ for $\overline{r}=1$:
improving the robustness of the best classifier decreases the room for improvement of the ensemble. 
This is a key finding: ensembling can do little to boost the robustness of a set of already robust classifiers.
We illustrate this in \Cref{fig:gaps_illustrations}a: for 1000 random classifiers, we show the gap $r^g_{c_B}$ vs $\overline{r}$ for the weights $\alpha_j$ that maximize $r^g_{c_B}$ for the specific ensemble.
The margin of improvement via ensembling is the gap between the diagonal and the bottom boundary of the orange region and indeed decreases to 0 as $\overline{r}{\to} 1$.

\stress{In practice, the prediction gap gains are likely even smaller.}
Most ensembles of random classifiers stay far from the bound and have even lower ensemble gap gain $r_{c_B}^g$ than \cref{eq:gap_gain_upper_bound} predicts, as \cref{fig:gaps_illustrations}a shows.
Furthermore, in reality, one has to pick a single set of weights $\alpha_j$ for all inputs $x$.
Often that is the uniform ensemble weight, \ie, $\alpha_j {=} \nicefrac{1}{N}$.
We show the gap gain for random classifiers with uniform weights in \cref{fig:gaps_illustrations}b.
Only a handful of ensembles remain in the \rone\ regime (above the diagonal in \cref{fig:gaps_illustrations}b) under uniform weights.
The majority of the points have $r^g_{c_B} {<} \overline r$ and are in \rtwo\ or \rthree\ (under the diagonal).
Therefore, in practice, ensembling rarely results in gap gains which is at odds with the \emph{ensembling for robustness} paradigm.
This is also true for real-world ensembles (see \cref{sec:experiments}).

\stress{Regime \Rthree\ is possible.}
\cref{fig:gaps_illustrations}b compares $r^g_{c_B}$ against $\overline r$, \ie, the most robust individual classifier. 
However, at different inputs $x$ the best classifier may be different.
Even if $g$ is always marginally less robust than the most robust classifier at a single $x$, $g$ might still be overall more robust than any single $f^j$.
To this end, \cref{fig:gaps_illustrations}c shows the ensemble gap $r^g_{c_B}$ against the \emph{worst} individual gap $\underline r$.
This shows that roughly half of the points are in gap regime \rthree, indicating that ensemble are often \emph{less robust than the least robust individual classifier}.
For \ucontstar\ ensembles this directly implies regime \Rthree. 
The same findings hold for the real-world classifiers in \Cref{sec:experiments}: for all of them the constituent models are on average more robust than the ensemble.

\stress{Ensembles can result in zero robustness.}
To make matters worse, not only is it possible that $r^g_{c_B}$ is smaller than all individual gaps, but it can even be 0, \ie, $Q_g = \{0\}$.
\begin{theoremEnd}[restate]{proposition}
    \label{thm:damning_ensembles_exist}
    For any set of $N\geq 2$ classifiers satisfying \diffca, there exist weights $\alpha_j$ for which the resulting ensemble has $r^g_{c_B} = 0$ and a certified perturbation set $Q_{g} = \{0\}$.
\end{theoremEnd}
\begin{proofEnd}
    First, note that $Q_{g(\alpha)}=\{0\}$ if $r^g_{c_B} = 0$, that is if the top two classes of $g$ have the same confidence.
    In other words, if $x$ is on the decision boundary for the ensemble $g$.
    Therefore, we want to show that it is possible to construct an ensemble for which the decision boundary passes through $x$.

    Let's first consider the case with two classifiers ($N=2$) and when the top prediction of $g$ is one of the top predictions of the individual classifiers for any $\alpha$: $c_A^g\in\{c_A^1, c_A^2\}, \forall \alpha\in[0,1]$.
    We denote by $g(\alpha)$ the ensemble $g(\alpha)=\alpha f_1 + (1-\alpha) f_2$.
    Therefore, we have $c_A^{g(\alpha)}=c_A^2$ when $\alpha$ is close to 0 and $c_A^{g(\alpha)}=c_A^1$ when $\alpha$ is close to 1.
    The switch between the two values happens at 
    \begin{equation}
        \alpha^\star = \frac{f^2_{c_A^2} - f^2_{c_A^1} }{f^1_{c_A^1} - f^1_{c_A^2} + f^2_{c_A^2} - f^2_{c_A^1}},
        \label{eq:damning_ensembles_exist_astar}
    \end{equation}
    if the denominator is not 0.
    It follows that when $\alpha=\alpha^\star$ we have that $g_{c_A^g} = g_{c_B^g}$, hence $r^g_{c_B}=0$ and $Q_{g(\alpha)}=\{0\}$.
    Note that if the denominator in \cref{eq:damning_ensembles_exist_astar} is 0, then $Q_{{g(\alpha)}}=\{0\}$ for all $\alpha$.
    
    Now consider the case when for some $\alpha$ the top prediction of $g$ is not one of the top predictions of the individual classifiers.
    Then we can split the domain $[0,1]$ for $\alpha$ into a subset that has only two top predictions and apply the above analysis to this subset.
    Therefore, when $N=2$ the proposition holds.

    To see that it holds for $N>2$, note that we can always fix $N-2$ of the $\alpha_j$ weights to 0.
    As long as we select two individual classifiers with different top predictions to have non-zero weights, we can apply the $N=2$ analysis to them.
    Therefore, the proposition holds for all $N$.
 \end{proofEnd}

\cref{fig:regime_1_example}c shows an example of $r_{c_B}^g=0$.
Therefore, ensembling not only can reduce robustness but can also result in an entirely non-robust classifier.
\Cref{fig:cifar10_hists,fig:imagenet_hists} show examples of this scenario occurring in practice.

\stress{Same top predictions prevent gap regime \rthree.}
The possibility of \rthree\ and the complete loss of robustness is certainly disappointing.
However, there is a simple way to prevent \rthree\ from occurring.
\cref{thm:damning_ensembles_exist} constructs an ensemble which has a decision boundary passing through $x$.
This is only possible if there are two classifiers in the ensemble with different top predictions (\diffca).
As long as all classifiers have the same top prediction, the ensemble cannot have a decision boundary passing through $x$.
Not only that, but also it will never be in regime \rthree, as illustrated by the red subset of ensembles in \cref{fig:gaps_illustrations}c.
\begin{theoremEnd}[restate,end]{proposition}
    \label{thm:top_same_never_worse_than_worst}
    No ensemble of $N$ classifiers over $K$ classes with $r^i \geq 0, i{=}1,\ldots,N$ satisfying \sameca\ can be in regime \rthree. 
\end{theoremEnd}
\begin{proofEnd}
    We have $N$ normalized $K$-class classifiers, so $f_j\in\RR^K$, $\sum_i f^j_i = 1$, $f^j_i >0$, for all $j\in 1,\ldots,N$.
    Furthermore, $g$ being a linear combination of $f^1,\ldots,f^N$ means that we have $g=\sum_{j=1}^N \alpha_j f_j$, for some $\alpha_j{>}0, \sum_j \alpha_j = 1$.
    $c_A^j$ and $c_B^j$ are the top two classes of $f_j$ and similarly $c_A^g$ and $c_B^g$ are the top two classes of $g$.
    We also have $r^j=f^j_{c_A^j}-f^j_{c_B^j}$, $r^g=g_{c_A^g}-g_{c_B^g}$, and $\underline r = \min_{j\in 1,\ldots,N} r^j$.

    First, observe that if all $c_A^j$ are the same and are equal to $c_A$, then $c_A^g$ must also be the same, \ie\ $c_A^g=c_A$.
    Then, $f^j_{c_B^g} \leq f^j_{c_B^j}, \forall j$.
    From these two observations we have:
    \begin{align*}
        \sum_{j=1}^N \alpha_j f^j_{c_A^g} &= \sum_{j=1}^N \alpha_j f^j_{c_A^j} \\
        \sum_{j=1}^N \alpha_j f^j_{c_B^g}&\leq \sum_{j=1}^N \alpha_j f^j_{c_B^j}.
    \end{align*}
    Hence the prediction gap of $g$ can be lower-bounded as:
    \begin{align*}
        r^g &= \sum_{j=1}^N \alpha_j f^j_{c_A^g} - \sum_{j=1}^N \alpha_j f^j_{c_B^g} \\
        &\geq \sum_{j=1}^N \alpha_j f^j_{c_A^j} -\sum_{j=1}^N \alpha_j f^j_{c_B^j} \\
        &= \sum_{j=1}^N \alpha_j \left( f^j_{c_A^j} - f^j_{c_B^j} \right) \\
        &= \sum_{j=1}^N \alpha_j r^j \\
        &\geq \min_{j\in 1,\ldots,N} r^j = \underline r.
    \end{align*}
    As $r^g \geq \underline r \geq 0$, the ensemble must be in regimes \rone\ or \rtwo.
\end{proofEnd}

Therefore, a practical way to avoid ensembles that are less robust than the least robust individual classifier is to enforce that all classifiers have the same top prediction. 

\stress{Summary.} Restricting the ensemble to satisfy \sameca\ and \samecb\, leads to regime \rtwo; no gap gain nor gap loss (\cref{thm:same_ca_cb}).
Dropping both conditions enables regime \rone\ but also \rthree.
However, keeping only condition \sameca, prevents regime \rthree\ while keeping \rone\ and \rtwo\ possible (\cref{thm:top_same_never_worse_than_worst}). 
For robust classifiers, the best-case ensemble prediction gap gains are very small (\cref{thm:gap_gain_upper_bound}). 
Finally, for ensembles in the \ucontstar\ mode \rone,\rtwo\ and \rthree\ imply \Rone,\Rtwo\ and \Rthree, respectively.

\subsection{Ensemble Certification for Different \texorpdfstring{$\CalS^{j}$}{Sj}}
\label{sec:ensemble_general_S}
\cref{sec:ensemble_same_S} considered the \ucontstar\ case where the prediction gap regimes are enough to reason about the certification regimes \Rone,\Rtwo,\Rthree.
It this section, we drop the \ucontstar\ requirement and show how the same results hold for general ensembles.

\stress{Regimes \Rone\ and \Rthree\ are possible for general smoothness.}
This follows trivially from the examples in \Cref{fig:regime_1_example} as general \SLipness subsumes the \ucontstar\ case.
\Cref{thm:damning_ensembles_exist} applies too, meaning that ensembles of robust classifiers can have $Q_g=\{0\}$ regardless of their \SLipness.

\stress{Same top predictions prevent regime \Rthree.}
As this is a non-existence result, it does not follow directly from \cref{thm:top_same_never_worse_than_worst}.
We would have to take into account the interaction of the shape and size of the $\CalS$ sets and the prediction gaps $r$.
\begin{theoremEnd}[all end]{auxlemma}
    \label{thm:top_same_gaps_are_lincomb}
    For any ensemble of $N$ normalized $K$-class classifiers satisfying \sameca\ it holds that the $i$-th class prediction gap $r^g_i$ of the ensemble is the weighted sum of the gaps $r^j_i$ of the individual classifiers:
    $$ r^g_i = \sum_{j=1}^N \alpha_j r^j_i, ~\forall i=1,\ldots,K.$$
\end{theoremEnd}
\begin{proofEnd}
    \begin{align*}
        r^g_i 
        &= \sum_{j=1}^N \alpha_j f_j^{c_A^g} - \sum_{j=1}^N \alpha_j f^j_i \\
        &= \sum_{j=1}^N \alpha_j f^j_{c_A^j} - \sum_{j=1}^N \alpha_j f^j_i \\
        &= \sum_{j=1}^N \alpha_j \left( f^j_{c_A^j} - f^j_i \right) \\
        &= \sum_{j=1}^N \alpha_j r^j_i.
    \end{align*}
\end{proofEnd}
\begin{theoremEnd}[end, restate]{proposition}
    \label{thm:top_same_never_smaller_than_intersection}
    No ensemble of classifiers as in \cref{thm:slip_addition} satisfying \sameca\ can be in regime \Rthree.
\end{theoremEnd}
\begin{proofEnd}
    We will deal only with the $N=2$ case as $N\geq 2$ follows by induction.
    Furthermore, we will assume that $\alpha_2=1-\alpha_1$ for simplicity.
    This doesn't affect the proof as the $\alpha_1+\alpha_2$ scaling does not affect the certificate (\cref{thm:slip_scaling}).

    We prove by contradiction. We restrict to same $c_A$, and assume that we have $\alpha_1$, classifier outputs and $\CalS$ sets such that:
    \begin{equation}
        Q_g \subset Q_1 \cap Q_2, \label{eq:top_same_never_smaller_than_intersection_cond_first}
    \end{equation}
    where
    \begin{align*}
        Q_1 &= \bigcap_{i\neq c_A} \left( \CalS_{i\sm c_A}^{1}\right)^{r^1_i} \\
        Q_2 &= \bigcap_{i\neq c_A} \left( \CalS_{i\sm c_A}^{2}\right)^{r^2_i} \\
        Q_g &= \bigcap_{i\neq c_A} \left(\alpha_1 \CalS_{i\sm c_A}^{1} \oplus (1-\alpha_1) \CalS_{i\sm c_A}^{2} \right)^{r^g_i}.
    \end{align*}
    Hence, \cref{eq:top_same_never_smaller_than_intersection_cond_first} becomes 
    \begin{equation}
        \bigcap_{i\neq c_A} \left(\alpha_1 \CalS_{i\sm c_A}^{1} {\oplus} (1{\sm}\alpha_1) \CalS_{i\sm c_A}^{2} \right)^{r^g_i} \subset \bigcap_{j=1,2} \bigcap_{i\neq c_A} \left( \CalS_{i\sm c_A}^{j}\right)^{r^j_i}.
        \label{eq:top_same_never_smaller_than_intersection_condition}
    \end{equation}
    This implies that there must be a point $x$ in the right-hand side of \cref{eq:top_same_never_smaller_than_intersection_condition} that is not in the left-hand side.
    This $x$ must satisfy:
    $$ \sup_{t\in \CalS_{i\sm c_A}^{j}} t^\top x \leq r^j_i \text{ for all } j=1,2, ~i\neq c_A.$$
    For the left-hand side of \cref{eq:top_same_never_smaller_than_intersection_condition}, using \sameca\ and \cref{thm:top_same_gaps_are_lincomb} we have:
    \begin{align*}
        &\left(\alpha_1 \CalS_{i\sm c_A}^{1} \oplus (1-\alpha_1) \CalS_{i\sm c_A}^{2} \right)^{r^g_i} \\
        =~&\left(\alpha_1 \CalS_{i\sm c_A}^{1} \oplus (1-\alpha_1) \CalS_{i\sm c_A}^{2} \right)^{\alpha_1 r^1_i + (1-\alpha_1) r^2_i}.
    \end{align*}
    We can see that $x$ must be in this polar set:
    \begin{align*}
         &\sup_{\substack{t_1\in \CalS_{i\sm c_A}^{1}\\t_2\in \CalS_{i\sm c_A}^{2}}}
        \left( \alpha_1 t_1^\top x + (1-\alpha_1) t_2^\top x \right) \\
        =~&\alpha_1\sup_{t_1\in \CalS_{i\sm c_A}^{1}} t_1^\top x + (1-\alpha_1) \sup_{t_2\in \CalS_{i\sm c_A}^{2}} t_2^\top x \\
        \leq~&\alpha_1 r^1_i + (1-\alpha_1) r^2_i.
    \end{align*}
    As this holds for all $i\neq c_A$, $x$ must also be in the intersection and hence in $Q_g$.
    This is a contradiction of the assumption that $x$ is not in $Q_g$.
\end{proofEnd}
Therefore, \sameca\ is sufficient to ensure regimes \Rone\ or \Rtwo, and avoid \Rthree.
This improves on \cref{thm:same_ca_cb} as relaxing the \samecb\ and \ucont\ conditions enable regime \Rone, while still preventing \Rthree, and extends \cref{thm:top_same_never_worse_than_worst} to general ensembles.

\stress{The margin of improvement is still limited.}
\cref{thm:gap_gain_upper_bound} showed that even in regime \rone, the gap gain is limited.
A similar observation holds for the robustness gain of arbitrary $\mathcal{S}^{j}$.
To simplify the analysis, we assume \sameca\ holds.
This is a reasonable assumption as \sameca\ prevents \Rthree\ as per \cref{thm:top_same_never_smaller_than_intersection}.
We will also assume that all $\CalS$ are of the same shape, \eg, norms, though not necessarily of the same size\footnote{This is more general than the \ucontstar\ condition in \cref{sec:ensemble_same_S} which restricted the sizes to also be the same.}.
This allows us to work with scalar radii instead of sets.
\begin{theoremEnd}[end, restate]{proposition}
    \label{thm:balls_max_improvement_bound}
    Take two classifiers $f^1,f^2:\RR^d\to\RR^K$ satisfying \sameca.
    Further, assume that all $h^j_{i\sm k}=f^j_i-f^j_k$ are $\epsilon_{j, i\sm k}\mathcal B_\star$-Lipschitz for some closed convex symmetric set $\mathcal B_\star$.
    Then, the maximum improvement in the certified radius $R^g$ of $g$ relative to the larger one of $R^1$ and $R^2$ is

    \resizebox{\columnwidth}{!}{
    $\begin{aligned}
        R^g {\sm} \max \{R^1, R^2\} {\leq} \frac{1}{\min\{ M^1, M^2 \} } {-} 
         \frac{\min\{ r^1_{c_B^1}, r^2_{c_B^2} \}}{\min\{ M^1, M^2 \} {+} \Delta}, 
    \end{aligned}$
    }

    where we have defined $M^k$ as $\min_{i\neq c_A} \epsilon_{k,i\sm c_A}$ and
    $\Delta$ as $\max_{k=1,2} \max_{i\neq c_A} ( \epsilon_{k,i\sm c_A} {-} M^k).$
\end{theoremEnd}

\begin{proofEnd}
    We assume that the predictions of each classifier are normalized, \ie, $\sum_i f^j_i=1, ~\forall j=1,\ldots,N$.
    Because all difference smoothness sets $\CalS$ have the same shape $\mathcal B_\star$ and the shape is closed under scaling and Minkowski sum (\cref{thm:sum_of_norm_balls}), we can simply work with a certified radius rather than a certified set.
    Note, however, that the shape of the certified sets would be the dual of the shape of the smoothness, \ie,\ $\mathcal B$. 
    Therefore, from \cref{thm:polar_norm_balls} we have:
    \begin{align}
        Q_j &= \bigcap_{i\neq c_A} (\epsilon_{j,i\sm c_A} \mathcal B_\star )^{r^j_i} \nonumber \\
            &= \min_{i\neq c_A} \left\{ \frac{r^j_i}{\epsilon_{j,i\sm c_A}} \right\} \mathcal B \hspace{3em} \text{ for } j=1,2\\
           &= \min_{i\neq c_A} \left\{ R^j_i \right\} \mathcal B,  \nonumber\\
           &= R^j \nonumber \\
       Q_\cup &= Q_1 \cup Q_2  \nonumber \\
              &= \max\{ \min_{i\neq c_A} R^1_i, \min_{i\neq c_A} R^2_i \} \mathcal B \label{eq:same_shape_same_ca_Qu} \\
              &= \max\{ R^{1}, R^{2} \} \mathcal B \nonumber \\
              &= R^\cup \mathcal B_\star \nonumber \\
       Q_\cap &= Q_1 \cap Q_2  \nonumber \\
              &= \min\{ \min_{i\neq c_A} R^1_i, \min_{i\neq c_A} R^2_i \} \mathcal B \label{eq:same_shape_same_ca_Qn} \\
              &= \min\{ R^{1}, R^{2} \} \mathcal B \nonumber \\
              &= R^\cap \mathcal B \nonumber
    \end{align}
    Next, note that as we have the same top predictions $c_A$, according to \cref{thm:top_same_gaps_are_lincomb} it holds that $r^g_i = \alpha_1 r^1_i + \alpha_2 r^2_i$.
    Therefore, from \cref{thm:sum_of_norm_balls,thm:polar_norm_balls} we have
    \begin{align}
        Q_g &= \bigcap_{i\neq c_A} \left( \alpha_1 \epsilon_{1, i\sm c_A} \mathcal B_\star   \oplus \alpha_2 \epsilon_{2, i\sm c_A} \mathcal B_\star \right)^{\alpha_1 r^1_i + \alpha_2 r^2_i} \nonumber \\
            &= \bigcap_{i\neq c_A} \left( ( \alpha_1 \epsilon_{1, i\sm c_A} + \alpha_2 \epsilon_{2, i\sm c_A} ) \mathcal B_\star \right)^{\alpha_1 r^1_i + \alpha_2 r^2_i} \nonumber \\
            &= \bigcap_{i\neq c_A}  \frac{\alpha_1 r^1_i + \alpha_2 r^2_i}{\alpha_1 \epsilon_{1, i\sm c_A} + \alpha_2 \epsilon_{2, i\sm c_A}} \mathcal B\label{eq:same_shape_same_ca_frac} \\
        &= \bigcap_{i\neq c_A}  R^g_i \mathcal B \nonumber \\
        &= \min_{i\neq c_A} \{ R_i^g \} \mathcal B\nonumber \\
        &= R^g \mathcal B. \nonumber
    \end{align}
    From \cref{eq:same_shape_same_ca_frac} it follows that the absolute values of $\alpha_1$ and $\alpha_2$ do not matter, only their relative size.
    This follows from the fact that $R^g_i$ is unchanged if we normalize $\alpha_1$ and $\alpha_2$ by diving by $\alpha_1+\alpha_2$.
    Therefore, with no loss of generality we will assume that $\alpha_1=\alpha$ and $\alpha_2=1-\alpha$ for the rest of the proof.
    This also follows from \cref{thm:slip_scaling}.

    While the claims of this proposition can be proven algebraically, we opt for a more intuitive graphical approach.
    First, notice that $R^g_i$ is a linear-fractional function in $\alpha$ and is monotonically increasing or decreasing from $R^2_i$ (for $\alpha=0$) to $R^1_i$ (for $\alpha=1$).
    Therefore, we can plot the $R^j_i$ and $R^g_i$ (as functions of $\alpha$) as in \cref{fig:same_shape_same_ca}.

    \begin{figure}
        \centering
        \includegraphics[width=0.75\columnwidth]{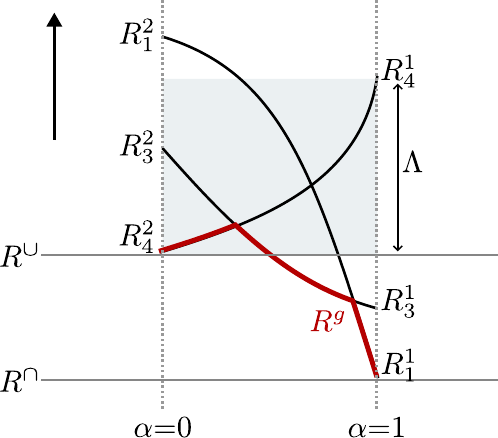}
        \caption{Illustration of the certified radii in \cref{thm:balls_max_improvement_bound}. The specific parameters used are $K=4, c_A=2$}
        \label{fig:same_shape_same_ca}
    \end{figure}
    
    From \cref{eq:same_shape_same_ca_Qu} we know that $R^\cup$ equals the larger one between the smallest radius on the left and the smallest radius on the right.
    Similarly, from \cref{eq:same_shape_same_ca_Qn} we have that $R^\cap$ is the smallest radius on either side.
    Finally, $R^g$ is the minimum of $R^g_i$ across $\alpha\in[0,1]$, or the thick red line in \cref{fig:same_shape_same_ca}.    
    
    Note the peak of $R^g$ cannot be larger than the smaller of the largest $R^1_i$ or the largest $R^2_i$ due to the monotonicity of $R^g_i$.
    Recall also that $\max \{R^1, R^2\} = R^\cup$ from \cref{eq:same_shape_same_ca_Qu}.
    Hence, $R^g - \max \{R^1, R^2\}$ is upper bounded by $\Lambda$, the height of the shaded region in \cref{fig:same_shape_same_ca}:
    \begin{equation}
        \Lambda {=} \min\{ \max_{i\neq c_A} R^1_i, \max_{i\neq c_A} R^2_i  \} \sm \max\{ \min_{i\neq c_A} R^1_i, \min_{i\neq c_A} R^2_i  \}
        \label{eq:lambda}
    \end{equation}

    As $r^j_{c_B^j}$ is the smallest gap for $f_j$ and as no gap can be larger than 1 we have:
    \begin{equation}
        \frac{r^j_{c_B^j}}{\epsilon_{j,i\sm c_A}} \leq R^j_i = \frac{r^j_i}{\epsilon_{j,i\sm c_A}} \leq \frac{1}{\epsilon_{j,i\sm c_A}}.
        \label{eq:Rij_bounds}
    \end{equation}
    Using this we can upper-bound the first term in \cref{eq:lambda}:
    \begin{equation}
        \begin{aligned}
            &\min \left\{ \max_{i\neq c_A} R^1_i, \max_{i\neq c_A} R^2_i \right\} \\
            \leq &\min \left\{ \max_{i\neq c_A} \frac{1}{\epsilon_{1,i\sm c_A}}, \max_{i\neq c_A} \frac{1}{\epsilon_{2,i\sm c_A}}  \right\}  \text{ (from Eq. \ref{eq:Rij_bounds})} \\
            = &\min \left\{ \frac{1}{M^1},  \frac{1}{M^2}  \right\} \\
            = &\frac{1}{\max\{ M^1, M^2 \}}.
        \end{aligned}
        \label{eq:lambda_left_bound}
    \end{equation}

    Using \Cref{eq:Rij_bounds} we can also lower-bound the second term in \cref{eq:lambda}:
    \begin{equation}
        \begin{aligned}
            &\max \left\{ \min_{i\neq c_A} R^1_i, \min_{i\neq c_A} R^2_i \right\} \\
            \geq &\max \left\{ \min_{i\neq c_A} \frac{r^1_{c_B^1}}{\epsilon_{1,i\sm c_A}}, \min_{i\neq c_A} \frac{r^2_{c_B^2}}{\epsilon_{2,i\sm c_A}}  \right\}  \\
            = &\max \left\{  \frac{r^1_{c_B^1}}{\max_{i\neq c_A} \epsilon_{1,i\sm c_A}}, \frac{r^2_{c_B^2}}{\max_{i\neq c_A} \epsilon_{2,i\sm c_A}}  \right\} \\
            \geq &\max \left\{  \frac{\min\{r^1_{c_B^1}, r^2_{c_B^2}\} }{\max_{i\neq c_A} \epsilon_{1,i\sm c_A}}, \frac{\min\{r^1_{c_B^1}, r^2_{c_B^2}\}}{\max_{i\neq c_A} \epsilon_{2,i\sm c_A}}  \right\} \\
            = &\frac{\min\{r^1_{c_B^1}, r^2_{c_B^2}\}}{\min\{ \max_{i\neq c_A} \epsilon_{1,i\sm c_A} , \max_{i\neq c_A} \epsilon_{2,i\sm c_A}\}} \\
            \geq &\frac{\min\{r^1_{c_B^1}, r^2_{c_B^2}\}}{\max\{ M^1, M^2 \} + \Delta }.
        \end{aligned}
        \label{eq:lambda_right_bound}
    \end{equation}
    Finally, substituting \cref{eq:lambda_left_bound,eq:lambda_right_bound} into \cref{eq:lambda}, we get the upper bound for $R^g - \max\{R^1, R^2\}$:
    \begin{align*}
        &R^g - \max\{R^1, R^2\} \\
        \leq &\Lambda \\
        \leq &\frac{1}{\max\{ M^1, M^2 \}} - \frac{\min\{r^1_{c_B^1}, r^2_{c_B^2}\}}{\max\{ M^1, M^2 \} + \Delta}.
    \end{align*} 
\end{proofEnd}

In the above proposition, $\min\{M^1, M^2\}$ refers to the radius of the least sensitive classifier, \ie, the one with smallest Lipschitz constant or \SLipness.
$\Delta$ measures how the Lipschitzness ranges amongst the classes and classifiers.
$\Delta=0$ implies that all $\epsilon_{k,i-c_A}$ are the same and therefore, all classifiers have the same Lipschitzness for all class pairs.
On the other hand, large $\Delta$ means that some classifiers are more robust for some class pairs while others are very sensitive for particular class pairs.

\cref{thm:balls_max_improvement_bound} is more restrictive when the individual classifiers have large predictions gaps ($r^1_{c_B^1}, r^2_{c_B^2}$) and/or similar Lipschitzness (small $\Delta$).
Both factors likely hold for robust classifiers: the large prediction gap is necessary for a large certificate and the similar Lipschitzness ensures similarly sized certificates for the different classes.
Therefore, in line with \cref{thm:gap_gain_upper_bound}, the ensembling improvement is only significant when the individual classifiers are not robust.

\stress{Sufficient conditions for improved certification are restrictive.}
Focusing again on the setting of \cref{thm:balls_max_improvement_bound}, we can provide sufficient conditions for regime \Rone:
\stepcounter{footnote}
\pgfmathparse{int(\value{footnote})}
\setvalue{sufficient_conditions_footnote = \pgfmathresult}
\footnotetext[\value{footnote}]{``Low confidences'' is formally defined in the proof.}
\begin{theoremEnd}[restate, end]{proposition}
    \label{thm:improvement_conditions_same_shape}
    Take an ensemble as in \cref{thm:balls_max_improvement_bound}.
    Assume two different second top predictions and that classes that are not in the top two predictions of any individual classifier have low confidences\footnotemark[\getvalue{sufficient_conditions_footnote}].
    Then \Rone\ occurs when:
    \begin{equation*}
        f^1_{c_A} {>} f^1_{c_B^2} {+} r^2_{c_B^2} \frac{\epsilon_{1, c_B^2\sm c_A}}{\epsilon_{2, c_B^2\sm c_A}} 
        \ \text{ and } \
    f^2_{c_A} {>} f^2_{c_B^1} {+} r^1_{c_B^1} \frac{\epsilon_{2, c_B^1\sm c_A}}{\epsilon_{1, c_B^1\sm c_A}}.
    \end{equation*}
\end{theoremEnd}
\begin{proofEnd}
    We restrict ourselves to the \sameca\ and different $c_B$ setting as this is the regime that prevents \Rthree\ and allows for \Rone\ (\cref{thm:same_ca_cb,thm:top_same_never_smaller_than_intersection}).
    We ask the classes that are not in the top two predictions of any individual classifier to have low predictions in order for them to not compete for the top ensemble prediction.
    Formally:
    $$ f^i_c < f^j_{c_B^k}, \forall j,k \in \{1, 2\}, \forall c \notin \{c_A\} \cup \{ c_B^l : l=1,2\}. $$
    From the proof of \Cref{thm:balls_max_improvement_bound} and our small third predictions assumptions we have:
    $$ R^\cup = \max \left\{ \frac{r^1_{c_B^1}}{\epsilon_{1,c_B^1\sm c_A}}, \frac{r^2_{c_B^2}}{\epsilon_{2,c_B^2\sm c_A}} \right\}, $$
    and
    \begin{align*}
        & R^g \\
        = & \min_{i\neq c_A} \left\{ \frac{\alpha_1 r^1_i + \alpha_2 r^2_i } {\alpha_1 \epsilon_{1,i\sm c_A} + \alpha_2 \epsilon_{2,i\sm c_A}} \right\} \\
        = & \min \left\{ \frac{\alpha_1 r^1_{c_B^1} {+} \alpha_2 r^2_{c_B^1} } {\alpha_1 \epsilon_{1,{c_B^1}\sm c_A} {+} \alpha_2 \epsilon_{2,{c_B^1}\sm c_A}}, 
        \frac{\alpha_1 r^1_{c_B^2} {+} \alpha_2 r^2_{c_B^2} } {\alpha_1 \epsilon_{1,{c_B^2}\sm c_A} {+} \alpha_2 \epsilon_{2,{c_B^2}\sm c_A}} \right\}.
    \end{align*}
    Both terms in the above minimum are monotonic.
    Therefore, the only way that $R^g > R^\cup$ for some $\alpha_1, \alpha_2$ is that the first term is decreasing while the second is increasing.
    This happens when
    \begin{align*}
        \frac{r^1_{c_B^2}}{\epsilon_{1,c_B^2-c_A}} &> \frac{r^2_{c_B^2}}{\epsilon_{2,c_B^2-c_A}} \\
        \frac{r^2_{c_B^1}}{\epsilon_{2,c_B^1-c_A}} &> \frac{r^1_{c_B^1}}{\epsilon_{1,c_B^1-c_A}},
    \end{align*}
    which, when rearranged, results in the conditions in the proposition.
\end{proofEnd}
    
The conditions in \cref{thm:improvement_conditions_same_shape} are rather limiting: the second class predicted by $f^2$ should have low enough confidence by $f^1$ and vice versa.
This means that ensembling ends up being beneficial at a fixed $x$ if each classifier has a different second prediction and all other predictions are very close to 0.
Therefore, regime \Rone\ is unlikely to occur unless the classifiers are carefully regularized.
\citet{pang19improving} suggest encouraging diversity among the non-maximal predictions.
\cref{thm:improvement_conditions_same_shape} theoretically justifies this approach.

\stress{Summary.}
The findings from \cref{sec:ensemble_same_S} hold also without the \ucontstar\ assumption.
Namely, all three regimes \Rone,\Rtwo,\Rthree\ are possible, \sameca\ prevents \Rthree\ (\cref{thm:top_same_never_smaller_than_intersection}) and the best-case ensembling improvement is small for robust classifiers (\cref{thm:balls_max_improvement_bound}).
Furthermore, we provide sufficient conditions for \Rone\ but these are severely limiting (\cref{thm:improvement_conditions_same_shape}).

\section{Discussion}
\label{sec:discussion}

In this section, we provide some comments on the implications and limitations of our theoretical analysis.

\stress{The conditions preventing regime \Rthree\ also prevent accuracy gain for the ensemble.}
\Cref{thm:top_same_never_smaller_than_intersection} showed that \sameca\ prevents regime \Rthree. 
However, ensembling cannot boost accuracy when in the \sameca\ regime.
Hence robustness seems to be at odds with accuracy, in line with the robustness-accuracy trade-off \citep{zhang_theoretically_2019,tsipras_robustness_2019}.

\begin{figure}
    \begin{minipage}[b]{0.35\columnwidth}
        \includegraphics[width=\textwidth]{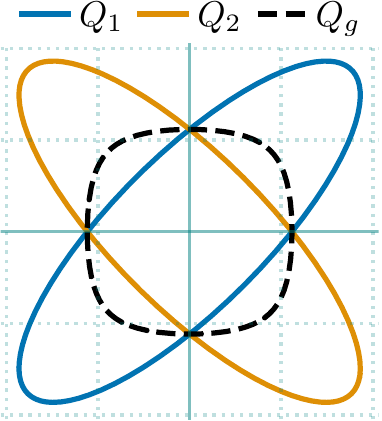}
    \end{minipage}%
    \hfill%
    \begin{minipage}[b]{0.60\columnwidth}
        \caption{Highly directional certificates can be ensembled to obtain directionally balanced certificates. $Q_g$ is constructed for $\alpha_j{=}\nicefrac{1}{2}$ and $r^1_{c_B}=r^2_{c_B}=1$.}
        \label{fig:aniso_to_iso}
    \end{minipage}%
    \vspace{-0.5cm}
\end{figure}

\stress{Ensembling can generate directionally-balanced certificates.}
When we have different shapes for $\CalS^{1}$ and $\CalS^{2}$, an ensemble can be used to trade-off classifiers that specialize in robustness in particular directions. 
As shown in \cref{fig:aniso_to_iso}, this technique can be used to construct more directionally-balanced certificates.
Therefore, depending on the notion of robustness, \Rtwo\ can be desirable when proper care is taken. 

\stress{The prediction gap and \SLipness are not independent.} 
Throughout this paper, we treated the \SLipness and the prediction gaps as two independent tools for boosting robustness. 
Intuitively, one would like to have as much as possible from both: smooth classifiers with high prediction gaps.
However, this is not possible.
The smoother a classifier is, the lower its prediction gaps are likely to be.
Therefore, the robustness gains from ensembling are likely even smaller than the already conservative bounds we have.
\Cref{sec:experiments} offers experiments demonstrating this effect.

\stress{Robustness over distributions rather than single points.}
\Cref{sec:smooth_ensembles} focused on point-wise robustness: all the results presented there are for a fixed $x$.
In reality, we are usually interested in the expected robustness over a distribution of inputs.
Even if the ensemble performs worse than the best individual classifier (\eg, \Rtwo) at all $x$, it might still be overall more robust than any individual classifier. 
Furthermore, the unfavourable conditions in \cref{thm:damning_ensembles_exist} might exist for some $x$, but it is likely that they are rare for real classifiers and distributions. 
We provide experimental observations to this effect in \cref{sec:experiments}.
The highlight is that for all ensembles considered, the ensemble certificates are smaller than these of the individual classifiers for more than 50\% of the inputs. 
Hence, real world ensembles seem to worsen robustness across distributions of inputs. 

\stress{Limitations of the \SLipness analysis.}
Most of the results in this paper are valid within the context of \Scerts: inferring certificates for ensembles from the \SLipness properties of the individual classifiers.
While this framework was necessary for the theoretical analysis, it might be conservative. 
Methods that construct certificates without direct reliance on ($\CalS$)-Lipschitzness properties, \eg, abstract interpretation \citep{gehr2018ai2} or SMT solvers \citep{huang2017safety}, might be able to provide larger certificates than what our theory predicts.
However, these methods cannot provide general theoretical analysis of the type we offer in this work.

\stress{Tightening via local \SLipness}
In \cref{thm:Slip_certificate,thm:Slip_certificate_diff}, we required that $f_i$ is $\CalS_i$-Lipschitz.
However, we do not necessarily need to constrain the \SLipness across the whole domain. 
Instead, we can work with $f_i$ locally $\CalS_i$-Lipschitz in a set $P$ containing $x$ \citep{weng2018evaluating}.
Note that when using local \SLipness, the certificate is valid only within $P$, \ie, the valid certificate is $P\cap Q$.

\section{Conclusion}
\label{conclusion}
We propose a new notion of Lipschitz continuity, namely $\mathcal{S}$-Lipschitzness, that offers tighter robustness certificates. 
We use this new framework to analyse the robustness properties of ensembles of classifiers.
Our results theoretically suggest that ensembling can improve the certification over the most robust individual classifier only under very strict conditions.
Moreover, even when improvements are possible, they are theoretically very small.
In addition, we prove that ensembling, if not done appropriately, can result in an ensemble worse than the least robust constituent classifier.
Even worse, it may result in a classifier with zero robustness.
Our theory suggests that boosting robustness via ensembling requires all classifiers to have the same top predictions and diverse second top predictions.

\section*{Acknowledgements}
AP has received funding from Toyota Motor Europe. 
FE and PT have received funding from FiveAI.
AS acknowledges partial support from the ETH AI Center fellowship.
AB has received funding from the Amazon Research Awards.
This work is supported by a UKRI grant Turing AI Fellowship (EP/W002981/1) and the EPSRC Centre for Doctoral Training in Autonomous Intelligent Machines and Systems (EP/S024050/1). 
AB would like to thank \href{https://www.konstmish.com/}{Konstantin Mishchenko} for the early insightful discussions. We also thank the Royal Academy of Engineering and FiveAI.

\newpage
\bibliography{bibliography}
\bibliographystyle{icml2023}

\clearpage
\newpage
\appendix
\counterwithin{figure}{section}
\renewcommand{\theHfigure}{figure.section.\thesection.\thefigure}

\onecolumn
\section{List of symbols}
\label{sec:symbols_list}

For the ease of the reader, we have summarized the notation used in the paper in the following table:

\renewcommand\arraystretch{1.25}
\begin{center}
\begin{tabular}{ll}
    $\alpha_j$ & The weight of the $j$-th classifier in the ensemble \\
    $\mathcal B$ & A norm ball \\
    $\mathcal B_\star$ & A dual norm ball \\
    $c_A^j$ & The class predicted  by the $j$-th classifier with the highest confidence\\
    $c_B^j$ & The class predicted  by the $j$-th classifier with the second highest confidence\\
    $c_A^g$ & The class predicted  by the ensemble with the highest confidence\\
    $c_B^g$ & The class predicted  by the ensemble with the second highest confidence\\
    \sameca & All top predictions in the ensemble are the same \\
    \diffca & At least two classifiers in the ensemble differ in their top prediction \\
    \samecb & All second highest predictions in the ensemble are the same \\
    $f$ & A classifier \\
    $f_i$ & The confidence for the $i$-th class of the classifier $f$ \\
    $f^j$ & The $j$-th classifier in the ensemble of classifiers \\
    $g$ & An ensemble of classifiers $f_1, \ldots, f_N$ \\
    $h_{i\sm k}$ & The difference of the confidence of classes $i$ and $k$ \\
    $i$ & Class index \\
    $j$ & Classifier index in an ensemble \\
    $K$ & Number of classes \\
    $L_i$ & The Lipschitz constant for the $i$-th class \\
    $N$ & Number of classifiers in the ensemble \\
    $Q$ & Certificate \\
    $Q_j$ & Certificate for the $j$-th classifier in the ensemble \\
    $Q_g$ & Certificate for the ensemble \\
    $r_i^j$ & The confidence gap between the top class and the $i$-th class for the $j$-th classifier in the ensemble \\
    $r_i^g$ & The confidence gap between the top class and the $i$-th class for the ensemble \\
    $\overline{r}$ & The maximum confidence gap in the ensemble ($\max_j r_{c_B}^j$) \\
    $\underline{r}$ & The minimum confidence gap in the ensemble ($\min_j r_{c_B}^j$) \\
    $R^j$ & Certified radius for the $j$-th classifier in the ensemble when $Q_j$ is a norm ball \\
    $R^g$ & Certified radius for the ensemble when $Q_g$ is a norm ball \\
    $\rho_\CalS$ & Support function \\
    $\CalS$ & Range space of gradients \\
    $\CalS_i$ & Range space of gradients for the $i$-th class \\
    $\CalS^j$ & Range space of gradients for the $j$-th classifier in the ensemble \\
    $\CalS_{i\sm k}$ & Range space of gradients for the difference of the confidence of classes $i$ and $k$ ($h_{i\sm k})$ \\
    $(\CalS)^r$ & Polar set of $\CalS$ with radius $r$ \\
    $\sigma$ & Smoothing Gaussian noise for randomised smoothing \\
    \ucont & Uniform continuity regime \\
    \ucontstar & Uniform continuity regime with all classifiers having the same \SLip for all classes \\
    \cwcont & Class-wise continuity regime \\
    \cdcont & Class-difference continuity regime \\
\end{tabular}
\end{center}

\twocolumn

\section{Experiments}
\label{sec:experiments}

In this appendix we describe several experiments that validate and illustrate the observations in the main body of the paper.

\stress{Experimental setup}
We use the ensembles trained by \citet{horvath2021boosting} that they have released publicly\footnote{Trained models are available at \url{https://github.com/eth-sri/smoothing-ensembles}}.
The classifiers are based on the ResNet20 and ResNet50 architectures \citep{he_deep_2016} and are trained respectively on CIFAR10 \citep{krizhevsky2009learning} and ImageNet \citep{russakovsky_imagenet_2015}.
We use randomized smoothing \citep{lecuyer2019certified,cohen_certied_2019} to obtain individual classifiers with known continuity properties ($\CalS$).
Concretely, a model smoothed with independent Gaussian noise with variance $\sigma^2$ is $\sqrt{\nicefrac{2}{\pi\sigma^2}}$-Lipschitz for the $\ell_2$ norm \citep{salman2019provably}.
As standard with randomized smoothing, each classifier is trained with Gaussian noise with variance matching the smoothing variance \citep{lecuyer2019certified}.

We consider the following ensembles:
\begin{enumerate}
\item Ensemble of $N{=}6$ ResNet20 classifiers trained on CIFAR10 ($K{=}10$), trained and smoothed with $\sigma{=}0.25$.
\item Ensemble of $N{=}6$ ResNet20 classifiers trained on CIFAR10 ($K{=}10$), trained and smoothed with $\sigma{=}0.50$.
    \item Ensemble of $N{=}6$ ResNet20 classifiers trained on CIFAR10 ($K{=}10$), trained and smoothed with $\sigma{=}1.00$.
    \item Ensemble of $N{=}3$ ResNet50 classifiers trained on ImageNet ($K{=}1000$), trained and smoothed with $\sigma{=}1.00$.
\end{enumerate}
We construct each ensemble with uniform weights $\alpha_j=\nicefrac{1}{N}$.
As all classifiers comprising an ensemble have the same $\CalS$ and are in the uniform continuity regime (\ucont), they are also in the \ucontstar\ regime.
Hence, as discussed in \cref{sec:ensemble_same_S}, we can directly infer the robustness certificates from the prediction gaps alone.

Note that for the experiments in this appendix, we \emph{first smoothen the individual classifiers and then ensemble them}.
This is as to make sure that the individual classifiers are smooth.
This is opposite to the procedure suggested by \citet{horvath2021boosting} and \citet{yang2022on}.
They \emph{ensemble first and smoothen the ensemble second}.

\begin{figure*}
    \centering
    \includegraphics[width=\textwidth]{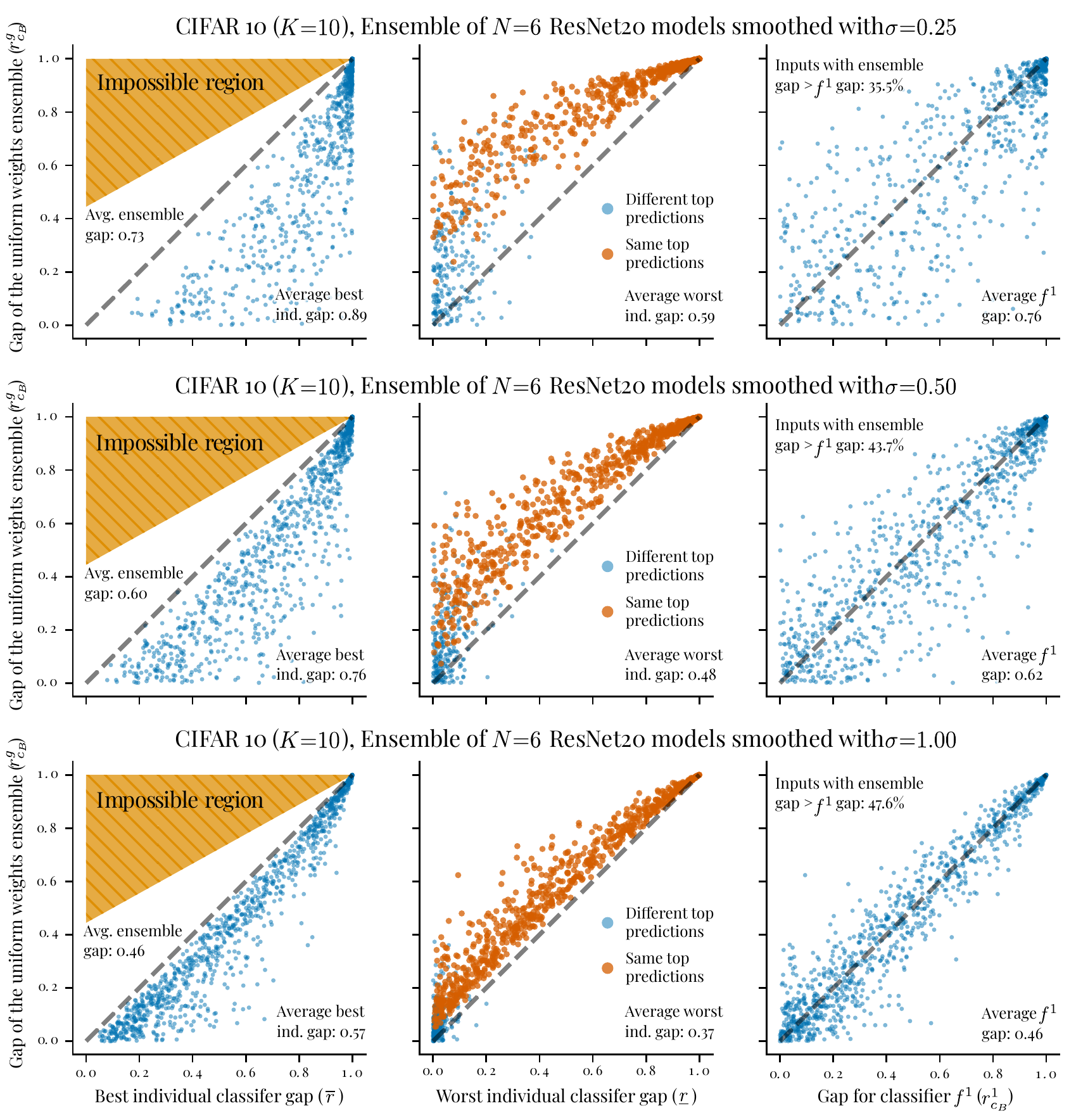}
    \caption{Gap of the uniform weights ensemble plotted against the best individual gap (left), the worst individual gap (center) and against the gap of one of the constituent classifiers (right). The plots against the other classifiers are similar and are hence omitted. 
    Each row shows one ensemble of 6 classifiers.
    Each individual classifier is a smoothed ResNet20 classifier trained by \citet{horvath2021boosting} using the train split of CIFAR10 and a different random seed.
    For these plots, we evaluate all classifiers at the same 1000 inputs from the CIFAR10 test split, each corresponding to a single point in the plots. 
    The impossible region in the leftmost plots follows from the bound from \cref{eq:gap_gain_upper_bound}.
    We have reported the average value for the horizontal and vertical axis for each plot. The percentage of inputs for which the ensemble has a larger gap than the individual classifier, is also shown in the rightmost plots.}
    \label{fig:cifar10_gaps}
\end{figure*}

\begin{figure*}
    \centering
    \includegraphics[width=\textwidth]{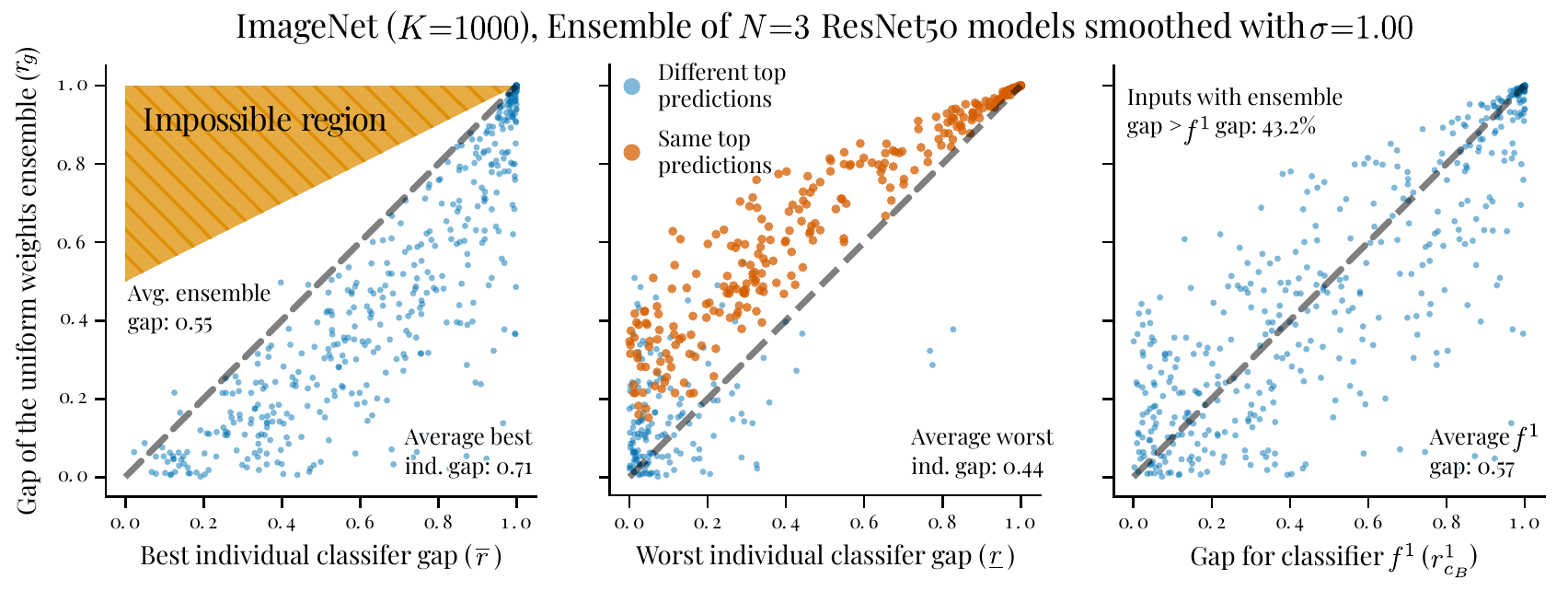}
    \caption{Gap of the uniform weights ensemble plotted against the best individual gap (left), the worst individual gap (center) and against the gap of one of the constituent classifiers (right). The plots against the other classifiers are similar and are hence omitted. 
    Each individual classifier is a smoothed ResNet50 classifier trained by \citet{horvath2021boosting} using the train split of ImageNet and a different random seed.
    For these plots, we evaluate all classifiers at the same 500 inputs from the ImageNet test split, each corresponding to a single point in the plots. 
    The impossible region in the leftmost plot follows from the bound from \cref{eq:gap_gain_upper_bound}.
    We have reported the average value for the horizontal and vertical axis for each plot. The percentage of inputs for which the ensemble has a larger gap than the individual classifier, is also shown in the rightmost plot.}
    \label{fig:imagenet_gaps}
\end{figure*}

\stress{Regime \Rone\ is possible but occurs rarely in practice.}
From the 1000 CIFAR10 inputs at which we evaluated the three ResNet20 ensembles  not a single one had an ensemble gap $r^g_{c_B}$ larger than the best individual classifier gap $\overline r$.
This is shown in the left-most column in \cref{fig:cifar10_gaps} that shows $r_{c_B}^g$ against $\overline{r}$: there is no points over the diagonal.
The ResNet50 ensemble, though, has 7 samples out of 500 in regime \Rone, \ie, for which the ensemble has a larger certified radius than the best individual classifier (left plot in \cref{fig:imagenet_gaps}).
However, this amounts to only 1.4\% of the inputs being in regime \Rone. 
Moreover, they are all very far from the bound on the maximum ensemble improvement from \cref{eq:gap_gain_upper_bound}.
This supports our hypothesis that, while the bound is achievable, the improvements ensembles would see in practice would be well below it.

\stress{Regime \Rthree\ occurs in practice but is also rare.}
Regime \Rthree, in which the ensemble fails to certify perturbations that every one of the individual classifiers certifies, does occur in practice as well.
This is evident from the points under the line in the middle plots in \cref{fig:cifar10_gaps,fig:imagenet_gaps} which show $r_{c_B}^g$ against $\underline{r}$.
For all four ensembles, there are inputs in regime \Rthree.
For the ResNet20 ensembles evaluated on CIFAR10, this occurs in respectively 3.9\%, 4.1\% and 3.3\% of the cases.
The ResNet50 ensemble has 10.2\% of its ImageNet samples in regime \Rthree.
These are much lower rates of occurrence than in the random ensemble in \cref{fig:gaps_illustrations} which is in regime \Rthree\ for 43.2\% of the inputs.
Still, all four ensembles have much larger rates of regime \Rthree\ compared to regime \Rone.
Therefore, this indicates that for real-world ensembles, most inputs are likely in regime \Rtwo, with some in regime \Rthree, and very few, if any, in regime \Rone.

\stress{Overall, the ensembles have smaller certificates than the individual classifiers.}
Most inputs of real-world ensembles seem to be in regime \Rtwo.
This means that the ensemble prediction gap  for an input $x$ (and hence certified radius) is between the smallest and the largest individual classifier gaps at $x$.
However, this does not tell us much about how the ensemble compares with a \emph{single} individual classifier, which is what one needs in order to decide whether it is better to use the ensemble or a single model.

We can make this comparison with the help of the leftmost and rightmost plots in \Cref{fig:cifar10_gaps,fig:imagenet_gaps} which show $r_{c_B}^g$ against respectively the best individual classifier gap $\overline{r}$ and the gap of one of the classifiers in the ensemble $r^1_{c_B}$.
The plots also show the average ensemble gap $r_{c_B}^g$ and average individual gap $r^1_{c_B}$ across all samples.
We can see that for all four ensembles, the average ensemble gap is smaller than the average gap of the individual classifier.
Therefore, as far as the average certified radius is concerned, the ensembles have \emph{lower} robustness than the individual classifier.
Furthermore, only between 35\% and 48\% of the inputs have an ensemble gap that is larger than the individual gap.
Hence, it appears that if one cares about robustness, they would be better off selecting one of the individual classifiers rather than the ensemble, for all four of these examples.

\stress{Ensembles of robust predictions can be non-robust in practice.}
\cref{thm:damning_ensembles_exist} showed that it is possible that ensembles which, at a given $x$, all have $r^j_{c_B}>0$, when ensembled can have $r^g_{c_B}=0$ and hence a certificate $Q_g=\{ 0\}$, regardless of the continuity properties of the classifiers.
One would hope that this is a purely theoretical curiosity and such situations do not occur in practice.
However, as all of the centre plots in \cref{fig:cifar10_gaps,fig:imagenet_gaps} show, for every ensemble, there are inputs for which the worst individual classifier has gap well above 0, while the ensemble gap is practically 0.
These are the points close to the horizontal axis.
We discuss two examples in more details.

\begin{figure*}
    \centering
    \includegraphics[scale=1.0]{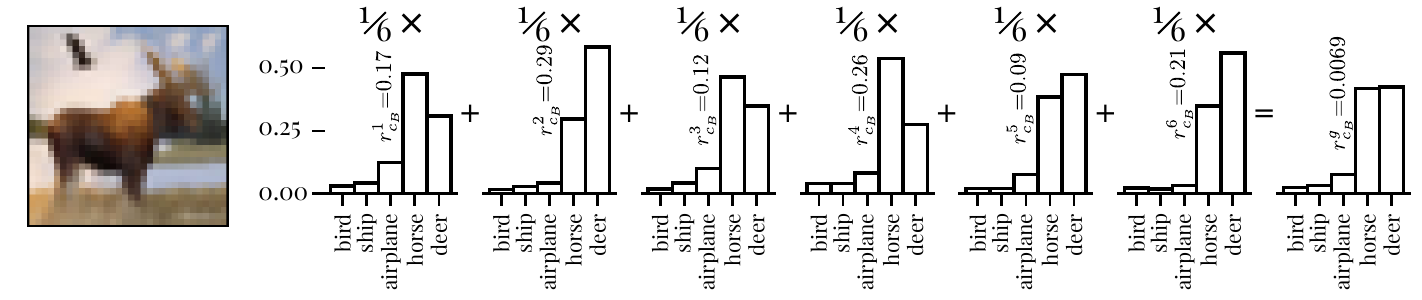}
    \caption{A CIFAR10 sample for which the ResNet20 ($\sigma=1.00$) ensemble is in regime \Rthree\ and has a certificate $Q_g$ barely larger than $\{0\}$. For clarity, only the 5 classes with the highest confidences are shown.}
    \label{fig:cifar10_hists}
\end{figure*}

\begin{figure*}
    \centering
    \includegraphics[scale=1.0]{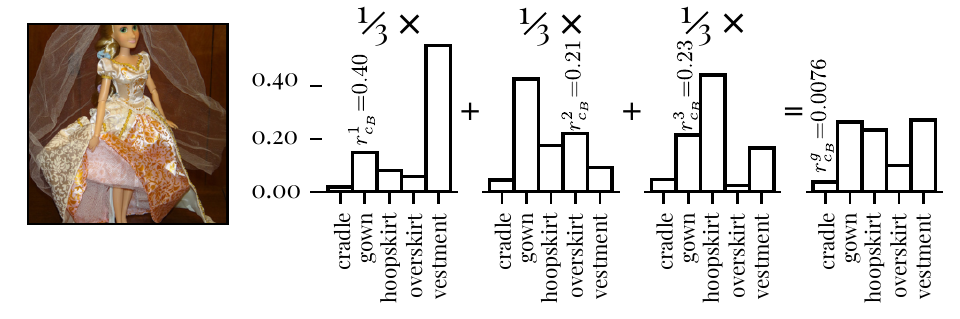}
    \caption{An ImageNet sample for which the ResNet50 ensemble is in regime \Rthree\ and has a certificate $Q_g$ barely larger than $\{0\}$. For clarity, only the 5 classes with the highest confidences are shown.
    }
    \label{fig:imagenet_hists}
\end{figure*}

\begin{figure*}
    \centering
    \includegraphics[width=\textwidth]{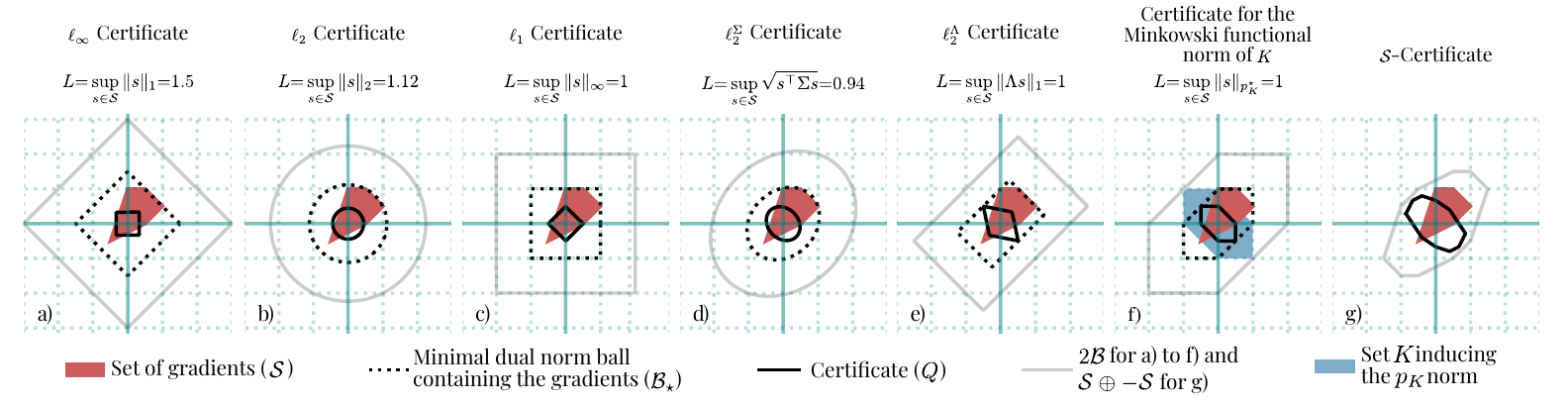}
    \caption{
        (\textbf{a}-\textbf{f}) are Lipschitz certificates for the set of gradients $\CalS=\{\nabla f_i(x) : x\in\RR^d, i=1,\ldots,K\}$. 
        We assume the \ucont\ mode and $r_{c_B}=1$. 
        $\mathcal B_\star$, the minimum dual norm ball containing $\CalS$, is shown. 
        The certificate $Q$ is the polar set $(2\mathcal B_\star)^1$. 
        For (\textbf{d}) and (\textbf{e}) we have  $\Sigma=\Lambda=\left[\begin{smallmatrix}\nicefrac{5}{4}&\nicefrac{1}{4}\\\nicefrac{1}{4}&\nicefrac{5}{4}\end{smallmatrix}\right]$. 
        (\textbf{f}) is the certificate constructed using the Minkowski functional norm (gauge) of $K$, the closed convex symmetric set marked in blue.
        (\textbf{g}) is the  \Scert for the same $\CalS$. 
        As there is no overapproximation of $\CalS$, the certificate is directly $Q=(\CalS\oplus-\CalS)^1$, the largest of them all.
        Note that (\textbf{a}) and (\textbf{g}) are the same as (\textbf{a}) and (\textbf{b}) in \cref{fig:linf_vs_S_cert}.
    }
    \label{fig:norms_certified_sets}
\end{figure*}

\Cref{fig:cifar10_hists} shows one CIFAR10 sample and its predictions by all 6 ResNet20 ($\sigma=1.00$) models and the ensemble prediction.
On average, the 6 classifiers have prediction gap $0.19$, with the smallest one being $\underline{r}=r^{5}_{c_B}=0.09$.
However, the ensemble gap is $r^g_{c_B}=0.0069$, more than an order of magnitude smaller than the smallest individual gap.
Hence, the ensemble certificate would too be more than an order of magnitude smaller than the smallest individual certificate.
This situation occurs as the 6 classifiers are split between classifying the input as a horse or a deer, resulting in very close predictions for the ensemble.

Similarly, the three ResNet50 classifiers have three different predictions for the input in \cref{fig:imagenet_hists}, none of which is the correct class (\texttt{overskirt}).
With $\underline r=0.21$ and $r^{g}_{c_B}=0.0076$, this leads to almost 30 times smaller certified radius of the ensemble compared with the least robust individual classifier.

In both of these examples, people would also likely be confused and would make mistakes.
Perturbing just a couple of pixels in the CIFAR10 input would likely be sufficient to nudge one in classifying the input as \texttt{horse} or as \texttt{deer}.
Therefore, lack of robustness in the ensemble might not be a bug, but in fact be a feature: a sign of better calibration.

\stress{Different top prediction is sufficient to ensure an ensemble is not in regime \Rtwo.}
From \cref{thm:top_same_never_worse_than_worst,thm:top_same_never_smaller_than_intersection}
we know that inputs for which all individual classifiers agree (\sameca) must be in regimes \Rone\ or \Rtwo.
From the center plots in \cref{fig:cifar10_gaps,fig:imagenet_gaps} one can observe that all inputs corresponding to this regime (in orange) are above the diagonal.
Therefore, our experimental results support \cref{thm:top_same_never_worse_than_worst,thm:top_same_never_smaller_than_intersection}.

\section{Additional examples}

\subsection{Examples of Lipschitz certificates for different norms}
\label{sec:additional_lipschitz_examples}

In the main text, we showed how to construct $\ell_p$ certificates (\cref{ex:lp_certificates}) and gave an illustration with an $\ell_\infty$ Lipschitz certificate in \cref{fig:linf_vs_S_cert}.
We offer some further examples here that we illustrate in \cref{fig:norms_certified_sets} using the same classifier as in \cref{fig:linf_vs_S_cert}.

\stress{Other $\ell_p$ certificates.}
Let's take a look at the other two commonly used $\ell_p$ certificates.
First, there is the $\ell_2$ certificate.
From \cref{ex:lp_certificates} and the H\"older inequality we have that the dual norm of $\ell_2$ is again $\ell_2$.
Hence, the certificate can be computed by finding the radius of the smallest $\ell_2$ ball that contains the gradients $\CalS$.
In the case illustrated in \cref{fig:norms_certified_sets}b we have $\sup_{s\in\mathcal S} \| s \|_2 {=} 1.12$.
Hence, $f$ is 1.12-Lipschitz with respect to the $\ell_2$ norm, and from \cref{thm:bounded_gradients_certificates} we have that the certificate $Q$ is $\{ \delta \in \RR^2 : \|\delta\|_2 \leq \nicefrac{1}{2.24}\}$ which corresponds to the circle marked with \solidline\ in \cref{fig:norms_certified_sets}b.

Similarly, the dual norm for $\ell_1$ is $\ell_\infty$.
Hence, we observe that $f$ is 1-Lipschitz with respect to the $\ell_1$ norm, that is $\sup_{s\in\mathcal S} \| s \|_\infty {=} 1$.
Therefore, the $\ell_1$ certificate is $Q=\{ \delta \in \RR^2 : \|\delta\|_1 \leq \nicefrac{1}{2}\}$, the rhombus marked with \solidline\ in \cref{fig:norms_certified_sets}c.

\stress{Anisotropic certificates.}
\Cref{thm:bounded_gradients_certificates} is not limited to $\ell_p$ norms.
Anisotropic certificates can be larger in some directions and smaller in others.
This is in contrast with the $\ell_p$ certificates which have the same radius in all directions.
This allows anisotropic certificates, in either of the \cdcont, \cwcont or \ucont modes, to be tighter in directions with smaller gradients.
For example, ellipsoidal certificates ---certificates with the $\ell^\Sigma_2$ norm defined as $\norm{\delta}^\Sigma_2 = \sqrt{ \delta^\top \Sigma^{-1} \delta}$--- can be constructed by bounding the gradients with its dual norm $\ell^{\Sigma^{-1}}_2$. %
Similarly, generalized cross-polytopes can be constructed with the $\ell^\Lambda_1$ norm defined as $\norm{\delta}^\Lambda_1 = \norm{ \Lambda^{-1} \delta}_1$ by bounding gradients with its dual norm $\ell^{\Lambda^{-1}}_\infty$.
The smallest norm balls (\dottedline) for $\Sigma=\Lambda=\left[\begin{smallmatrix}\nicefrac{5}{4}&\nicefrac{1}{4}\\\nicefrac{1}{4}&\nicefrac{5}{4}\end{smallmatrix}\right]$ and the corresponding certificates (\solidline) are shown in \cref{fig:norms_certified_sets}d and e.
Refer to \citet{eiras2022ancer} for further details.

\stress{Arbitrary norms defined as Minkowski functionals.}
Any closed convex symmetric set $K\subset\RR^d$ containing the origin gives rise to a norm on $\RR^d$ defined as $ p_{K}(x):=\inf\{a\in \RR :a>0{\text{ and }}x\in aK\}.$
This is called \emph{Minkowski functional} or \emph{gauge} of $K$ \citep{schechter1995handbook}.
Intuitively, $p_K(x)$ measures how much we need to scale $K$ in order to have $x$ barely fitting in it, \ie, $x$ being on the border of the scaled $K$.
\cref{fig:linf_vs_S_cert}f illustrates such a closed convex symmetric set $K$ in \bluepatch and the minimum dual $p_K^\star$ norm containing $\CalS$ (\dottedline) with a radius $\sup_{s\in\mathcal S} \| s \|_{p_K^\star} {=} 1$.
Therefore, the certificate is the $p_K$ norm ball of radius $\nicefrac{1}{2}$, shown in \solidline.

\stress{Comparison with the \Scert.}
The \Scert shown with \solidline in \Cref{fig:linf_vs_S_cert}g is the largest of all seven certificates.
\Cref{thm:scert_always_larger_than_lip} shows that this must always be the case: there is no norm for which the Lipschitz certificate will be a strict superset of the \Scert.
More detailed explanation is offered in \cref{sec:SLip_larger_than_Lip} in the main text.

\subsection{One-dimensional binary classifier example}
\label{ex:1d_linear_classifier}

\begin{figure}
    \includegraphics[width=0.8\columnwidth]{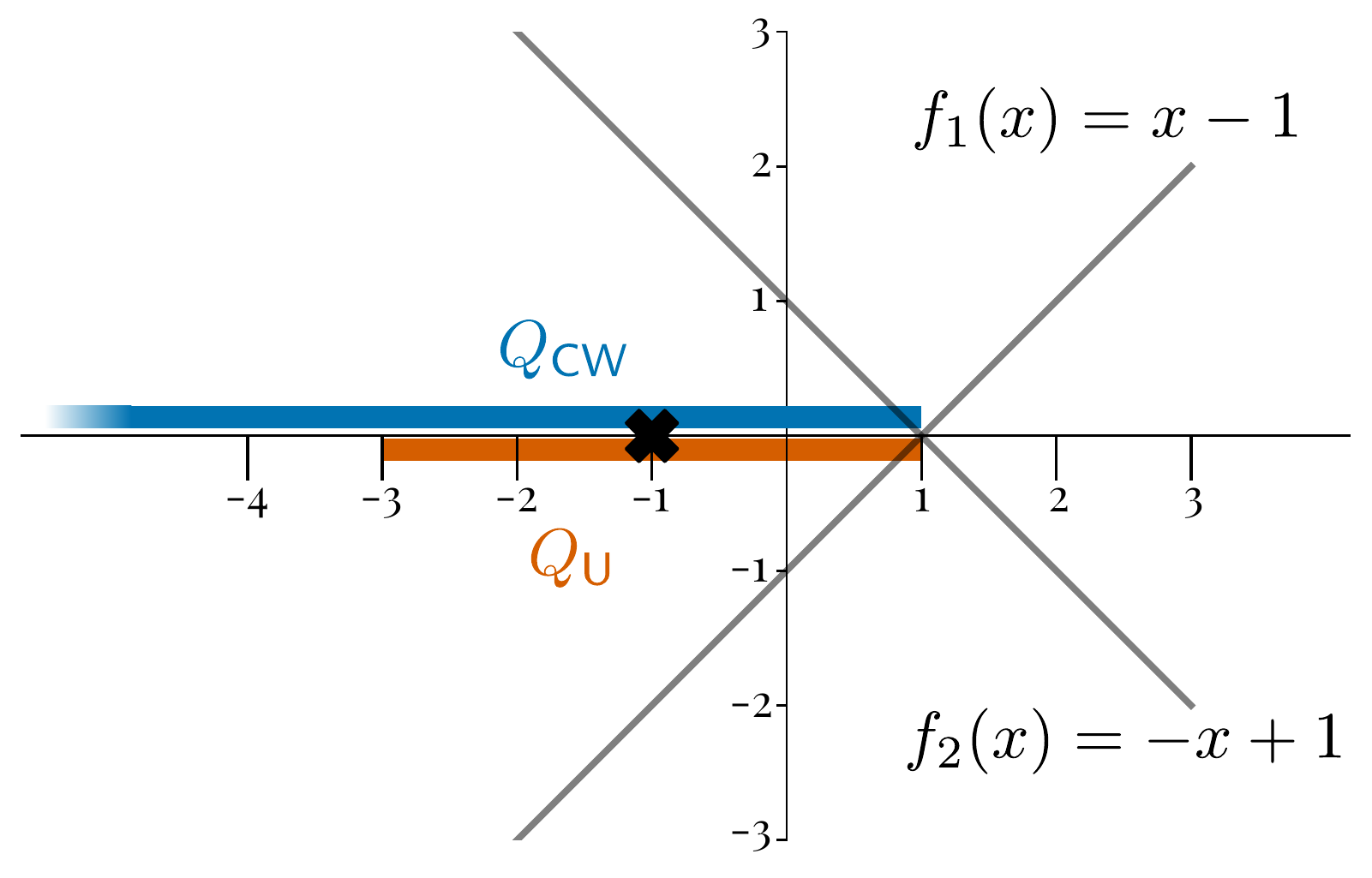}
    \caption{Illustration of the one-dimensional binary classifier example in \cref{ex:1d_linear_classifier}.}
    \label{fig:cert_ex_1D}
\end{figure}

Linear classifiers are easy to analyse as their $\CalS$ sets are singleton sets.
Let's then see the difference between the Lipschitz and the \SLip certificates for a one-dimensional linear binary classifier defined as
\begin{equation*}
    f_1(x) = x-1, \hspace{1cm}    f_2(x) = -x+1.
\end{equation*}

For this classifier we have $\CalS_1=\{+1\}$ and $\CalS_2=\{-1\}$.
We want to compute certificates for the input $x=-1$.
Hence $c_A=2$ and $r_{c_B}=f_2(-1)-f_1(-1)=4$.
Let's first consider the \cwcont\ certificate from \cref{eq:Slip_certificate_cw}.
We have $Q_\textsf{CW}=(\CalS_1\oplus-\CalS_2)^r=(\{1\}\oplus-\{-1\})^4=\{2\}^4=(-\infty, 2].$
This certificate is shown in blue in \Cref{fig:cert_ex_1D}.
If we instead construct the \ucont\ certificate by taking the smallest $\CalS$ such that $f_1$ and $f_2$ are \SLip, then we get $\CalS=\{-1, 1\}$.
Certifying using this $\CalS$, \cref{eq:Slip_certificate_uniform} gives us $Q_\textsf{U} = (\CalS\oplus-\CalS)^r = \{-2, 0, 2\}^4 = [-2, 2]$.
This certificate is shown in orange in \Cref{fig:cert_ex_1D}.
$f$ is 1-Lipschitz with respect to any $\ell_p$ norm and the Lipschitz certificate \cref{thm:bounded_gradients_certificates} results in the same certified perturbation set: $[-2, 2]$ for any $\ell_p$.
Therefore, even in this simple case, we see that the \cwcont\ \Scert covers the whole domain in which $f$ predicts 2 while the Lipschitz approach and the \ucont\ \Scert are limited to the largest \emph{symmetric} perturbation set.

\subsection{Derivation of the certificates in \cref{fig:2d_linear_classifier}}
\label{sec:2d_linear_classifier_workedout}

\begin{figure}
    \includegraphics[width=0.95\columnwidth]{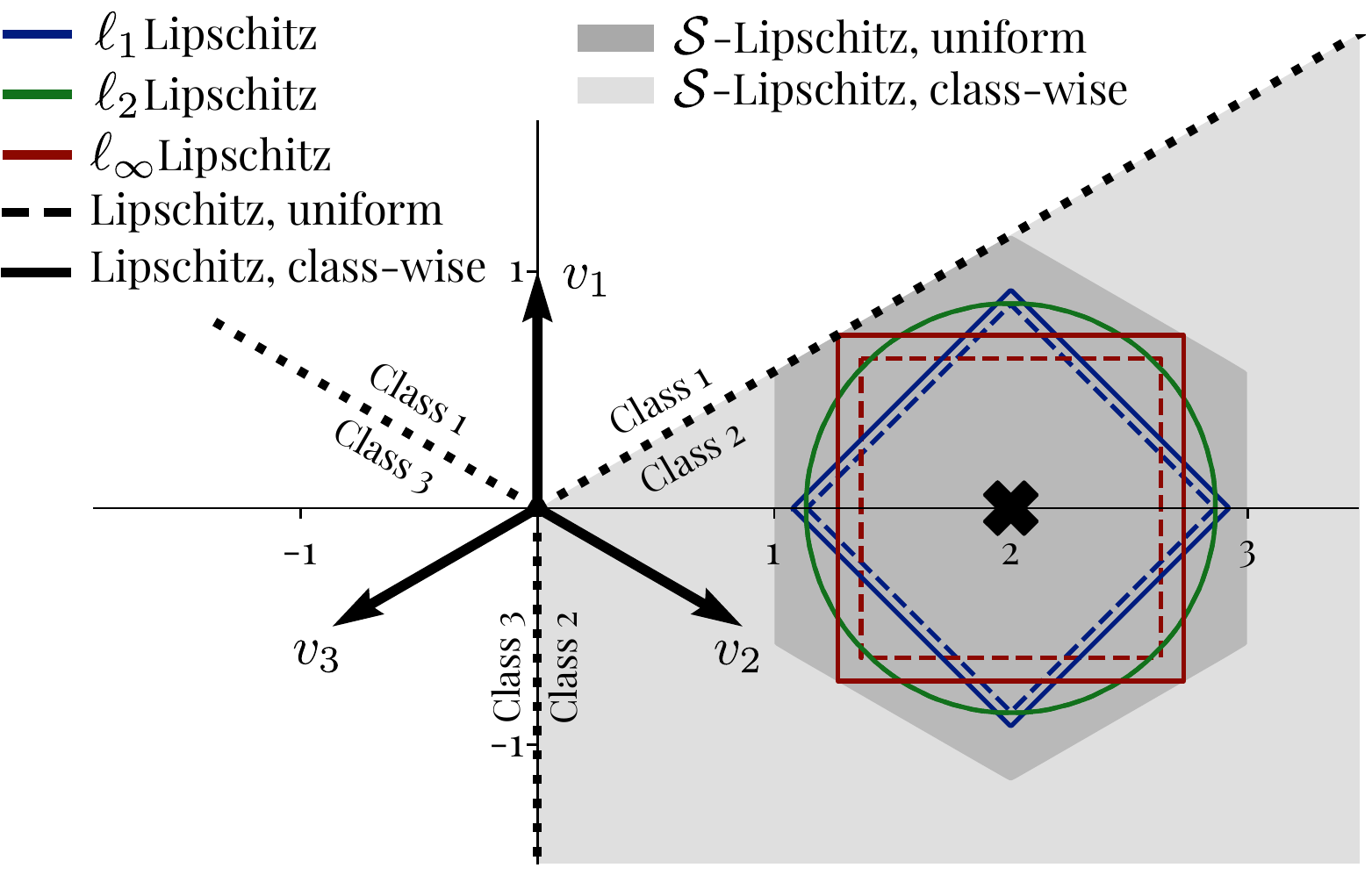}
    \caption{Illustration of the two-dimensional three-way classifier example from \cref{sec:2d_linear_classifier_workedout} and \cref{fig:2d_linear_classifier}.}
    \label{fig:cert_ex_2D_appendix}
\end{figure}

This is an extended explanation of \cref{fig:2d_linear_classifier} with all the intermediate steps and calculations.

Consider the 3-class two-dimensional linear classifier defined as:
\begin{align*}
    f_1(x) &= x^\top v_1 = [0, 1] \cdot x  \\
    f_2(x) &= x^\top v_2 = [\nicefrac{\sqrt{3}}{2}, -\nicefrac{1}{2}] \cdot x \\
    f_3(x) &= x^\top v_3 = [\nicefrac{-\sqrt{3}}{2}, -\nicefrac{1}{2}] \cdot x 
\end{align*}
We want to construct a certificate for $x_0=[2, 0]^\top$.
We then have $f_1(x_0)=0$, $f_2(x_0)=\sqrt{3}$, $f_3(x_0)=-\sqrt{3}$, $c_A=2$, $c_B=1$, $r_1=\sqrt{3}$, $r_3=2\sqrt{3}$.

Let's first consider the \ucont\ Lipschitz case using the observation that $f_1, f_2$, and $f_3$ are $L^p$-Lipschitz for the $\ell_p$ norm with $L^1=L^2=1$, $L^\infty=\nicefrac{(\sqrt{3}+1)}{2}$ (from \cref{thm:gradients_lipschitz}).
The respective certificates would be the $\ell_p$ ball with radius $\nicefrac{r_1}{2L^p}$, as shown in \cref{fig:cert_ex_2D_appendix}.
Now, let's compare with the \cwcont\ case.
\begin{align*}
    \CalS_1&{=}\left\{\begin{bmatrix}0\\1\end{bmatrix}\right\} 
    &L_1^1 &{=} 1              &L_1^2 &{=} 1     &L_1^\infty &{=} 1 \\
    \CalS_2&{=}\left\{\begin{bmatrix}\nicefrac{\sqrt{3}}{2}\\ -\nicefrac{1}{2}\end{bmatrix}\right\} 
    &L_2^1 &{=} \frac{\sqrt{3}}{2}     &L_2^2 &{=} 1     &L_2^\infty &{=} \frac{\sqrt{3}+1}{2} \\
    \CalS_3&{=}\left\{\begin{bmatrix}-\nicefrac{\sqrt{3}}{2}\\ -\nicefrac{1}{2}\end{bmatrix}\right\}
    &L_3^1 &{=} \frac{\sqrt{3}}{2}     &L_3^2 &{=} 1     &L_3^\infty &{=} \frac{\sqrt{3}+1}{2}
\end{align*}
The respective certificates would be the intersection of the $\ell_p$ balls with radius $\min\{\nicefrac{r_1}{(L_1^p+L_2^p)},\nicefrac{r_3}{(L_3^p+L_2^p)}\}$.
For $\ell_1$ and $\ell_\infty$ we observe increased certified radii when using \cwcont\ Lipschitzness: respectively from $\nicefrac{\sqrt{3}}{2}$ to $\nicefrac{2\sqrt{3}}{(2+\sqrt{3})}$ and from $\nicefrac{\sqrt{3}}{(1+\sqrt{3})}$ to $\nicefrac{2\sqrt{3}}{(3+\sqrt{3})}$.
The certified radius for $\ell_2$ remained unchanged: $\nicefrac{\sqrt{3}}{2}$: that's because $L_1^2=L_2^2=L_3^2$ and hence we don't overapproximate the true smoothness in the \ucont\ case. 

Next, let's do the same analysis using \SLipness instead.
In the \ucont\ case, we have that $f$ is \SLip with $\CalS=\CalS_1\cup\CalS_2\cup\CalS_3$.
Therefore, the certified set is the hexagon in \cref{fig:cert_ex_2D_appendix} (via \cref{thm:polar_as_hyperplane_intersection}).

Finally, let's take a look at the \cwcont\ \Scert: this should give us the largest certified region.
Again using \cref{thm:polar_as_hyperplane_intersection} we have 
\begin{align*}
    Q &=(\CalS_1\oplus-\CalS_2)^{r_1} \cap (\CalS_3\oplus-\CalS_2)^{r_2} \\
      &=\{ x\in\RR^d : [\nicefrac{-1}{2}, \nicefrac{\sqrt{3}}{2}] \cdot x \leq 1 ~\land~ [\nicefrac{-1}{2}, 0] \cdot x \leq 1 \}.
\end{align*}
This is all of the domain that $f$ classifies as class 2.

Hence, the \cwcont\ \SLip approach gives us the maximum possible certified domain: the whole preimage of the class 2 prediction.
All \cwcont\ Lipschitz certificates are smaller than the \cwcont\ \Scert as they consider only the norm of the gradients and ignores their orientation.
Similarly all \ucont\ Lipschitz certificates are smaller than the \ucont\ \Scert.
The \ucont\ \Scert is smaller than the \cwcont\ \Scert as it ignores the class-wise differences, and similarly the \ucont\ Lipschitz certificate is smaller than the \cwcont\ Lipschitz certificate.

\subsection{Example for \cref{thm:same_ca_cb}}
\label{ex:same_ca_cb_same_shape}

\begin{figure}
    \includegraphics[width=0.95\columnwidth]{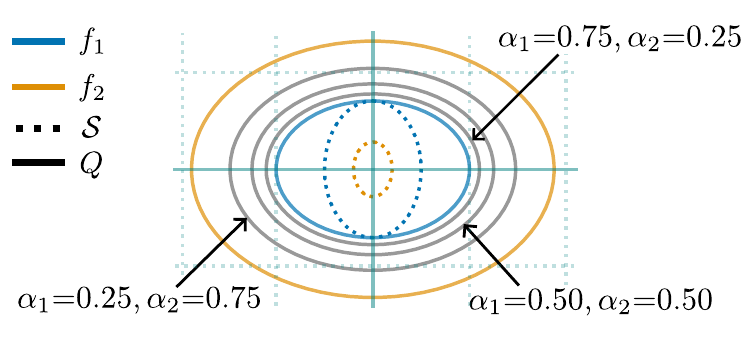}
    \caption{Illustration for the example in \cref{ex:same_ca_cb_same_shape}.}
    \label{fig:same_ca_cb_same_shape}
\end{figure}

Take two classifiers $f^1, f^2:\RR^2\to\RR^K$ under the conditions in \Cref{thm:same_ca_cb}.
Assume further that their \SLip sets have the same shape but possibly different sizes.
That is, $\CalS^{1}=\epsilon_1 \mathcal B_\star, \CalS^{2}=\epsilon_2 \mathcal B_\star$, $\epsilon_1, \epsilon_2>0$ where $\mathcal B_\star=\{x\in\RR^d~:~\norm{x}_\star\leq 1\}$ for some norm $\|\cdot\|_\star$.
We use $\mathcal{B}$ to denote the unit ball defined by the dual norm $\|\cdot\|$. %
From \cref{thm:Slip_certificate_diff}, we have
\begin{align*}
    Q_1 &{=} \frac{r_{c_B}^1}{2\epsilon_1} \mathcal B, 
        &Q_2 &{=} \frac{r_{c_B}^2}{2\epsilon_2} \mathcal B, 
        &Q_g  
      &{=}\frac{\alpha_1 r_{c_B}^1 + \alpha_2 r_{c_B}^2}{2(\alpha_1\epsilon_1 + \alpha_2 \epsilon_2)} \mathcal B.
\end{align*}
The radius of $Q_g$ interpolates from $\nicefrac{r_{c_B}^1}{2\epsilon_1}$ to $\nicefrac{r_{c_B}^2}{2\epsilon_2}$ and can never be larger than $\max\{ \nicefrac{r_{c_B}^1}{2\epsilon_1},\nicefrac{r_{c_B}^2}{2\epsilon_2}\}$.
Therefore, in this setting, ensembling will always result in a smaller certified radius than the most robust individual classifier.

We illustrate this phenomenon in \cref{fig:same_ca_cb_same_shape}.
Consider the anisotropic ellipsoidal norm $\norm{x} {=} \sqrt{x^\top \left[ \begin{smallmatrix} 1&0\\0&2\end{smallmatrix} \right] x}$ (see \cref{sec:additional_lipschitz_examples} for further details on this norm).
The radii of $\CalS^{1}$ and $\CalS^{2}$ are respectively $\epsilon_1{=}\nicefrac{1}{2}$ and $\epsilon_2{=}\nicefrac{1}{5}$ (shown in \dottedline), and their prediction gaps are $r_{c_B}^1{=}1$ and $r_{c_B}^2 {=}\nicefrac{3}{4}$. 
We show the certificate $Q_1$ for $f_1$ as the smallest ellipse and the certificate $Q_2$ for $f_2$ as the largest one.
We also show three sets of mixing coefficients $\alpha_1,\alpha_2$ in grey, which all fall between $Q_1$ and $Q_2$.
This illustrates how in the \ucont, \sameca, \samecb\ and same shape of the \SLipness regime, we will always have the largest certified radius by picking the best individual classifier ($f_2$ in this case), instead of ensembling.

\section{Deferred Proofs}

\printProofs

\end{document}